\documentclass{article}

\PassOptionsToPackage{numbers, compress}{natbib}

\usepackage[main, final]{neurips_2026}

\usepackage{hyperref}
\usepackage{pifont}
\usepackage{multirow}
\usepackage{graphicx}
\usepackage{float}
\usepackage[utf8]{inputenc} 
\usepackage[T1]{fontenc}    
\usepackage{url}            
\usepackage{booktabs}       
\usepackage{amsfonts}       
\usepackage{nicefrac}       
\usepackage{microtype}      
\usepackage{xcolor}         
\usepackage{enumitem}
\usepackage{pifont}
\usepackage{multirow}
\usepackage{makecell}
\usepackage{amsmath}
\usepackage{ulem}
\usepackage{wrapfig}
\usepackage{caption}
\usepackage{colortbl}
\usepackage{color}
\usepackage[skins, most]{tcolorbox}
\usepackage{fancyvrb}
\usepackage{listings}

\definecolor{lightgray}{gray}{0.9}

\newtcolorbox{findingbox}{
    enhanced,
    colback=gray!4!white,       
    colframe=black!65,          
    boxrule=0.8pt,
    arc=3pt,
    drop shadow=black!12,       
    left=8pt, right=8pt,
    top=6pt, bottom=6pt,
    width=\linewidth,
    boxsep=0pt
}

\title{BusterX: MLLM-Powered AI-Generated Video Forgery Detection and Explanation}

\author{
    \textbf{Haiquan Wen}$^{1}$\textsuperscript{$*$} \quad \textbf{Yiwei He}$^{1}$\textsuperscript{$*$} \quad \textbf{Zhenglin Huang}$^{1}$\textsuperscript{$*$} \quad \textbf{Tianxiao Li}$^{1}$ \quad \textbf{Zihan Yu}$^{1}$ \\ \textbf{Xingru Huang}$^{5}$ \quad \textbf{Lu Qi}$^{4}$ \quad \textbf{Baoyuan Wu}$^{3}$ \quad \textbf{Xiangtai Li}$^{2}$ \quad \textbf{Guangliang Cheng}$^{1}$\textsuperscript{$\dagger$} \\
    \\
    {\normalsize $^1$University of Liverpool \quad $^2$Nanyang Technological University} \\
    {\normalsize $^3$The Chinese University of Hong Kong, Shenzhen} \\
    {\normalsize $^4$Wuhan University \quad $^5$Hangzhou Dianzi University} \\
    {\normalsize \textsuperscript{$*$} Equal Contribution.}
    {\normalsize \textsuperscript{$\dagger$} Corresponding author: guangliang.cheng@liverpool.ac.uk} \\
    {\normalsize Project Page: \url{https://github.com/l8cv/BusterX}} \\
}

\begin{document}

\maketitle

\begin{abstract}
As generative video models become increasingly realistic, detecting AI-generated videos requires systems that offer both accuracy and interpretability. However, applying Multimodal Large Language Models (MLLMs) to video forensics is currently limited by outdated datasets, simplistic evaluation protocols, and a reliance on black-box classification. 
To address these issues, we introduce a comprehensive dataset, benchmark, and baseline model for video forgery detection. 
First, we present \textbf{GenBuster-200K}, a fair dataset of over 200,000 high-quality videos sourced from state-of-the-art generators, featuring diverse real-world scenarios. 
Second, we propose \textbf{GenBuster-Bench}, a diagnostic benchmark spanning three progressive tracks (In-Domain, Out-of-Domain, and In-the-Wild) to evaluate models across \textit{domain shifts} and \textit{generational shifts}.
Finally, we develop \textbf{BusterX}, a simple MLLM baseline with Reinforcement Learning (RL) training. Instead of direct binary classification, BusterX formulates detection as a visual reasoning task, where the generated reasoning chain serves as detector itself. Experimental results demonstrate that BusterX outperforms several leading MLLMs in both detection accuracy and rationale quality.

\end{abstract}

\section{Introduction}
\label{sec:intro}

Recent models, such as Seedance 2.0 \cite{seedance2.0}, enable users to synthesize highly realistic videos with minimal effort. While beneficial for creative applications, this accessibility also facilitates malicious uses, such as political misinformation and financial fraud. Consequently, there is a growing need for reliable and interpretable video forgery detection systems. To provide human-readable insights, recent research has started exploring Multimodal Large Language Models (MLLMs) as an alternative to traditional binary classifiers.

Despite their potential, applying MLLMs to video forensics is currently limited by the lack of high-quality data, simplistic evaluation protocols, and a reliance on direct classification. Specifically, we identify these critical limitations in the current research landscape:
\textbf{1) Outdated and Biased Data:} Existing video datasets \cite{bai2024ai, ma2024decof, chen2024demamba, ni2025genvidbench, DBLP:journals/tmm/JuJCGL24, khalid2021fakeavceleb} frequently rely on early-stage generators, producing easily recognizable artifacts. Furthermore, they often overlook demographic fairness, failing to account for gender and racial diversity.
\textbf{2) Aggregated and Black-Box Evaluation:} Current benchmarks predominantly treat video forgery detection as a simple binary classification task on an aggregated test set. This approach obscures a model's specific weaknesses---whether it fails to generalize to out-of-domain models or breaks down under real-world social media degradations. Crucially, they also ignore the \textit{explainability} of detection models. In forensic scenarios, providing a human-readable rationale is just as important as the binary verdict itself.
\textbf{3) Lack of Forensic Reasoning:} Recent methods that adapt MLLMs for detection either use them as auxiliary to assist binary classification, or merely to generate post-hoc explanations. Neither approach treats step-by-step physical and spatiotemporal reasoning as the detection mechanism itself.

\begin{table}[t]
    \centering
    \caption{Comparison with existing datasets. GenBuster outperforms prior works in scale, visual quality and fairness (balanced gender, age, and ethnicity distributions). In-the-wild refers to whether the dataset includes videos sourced from social media.}
    \small
    \resizebox{1\textwidth}{!}{
    \begin{tabular}{ccccccccccc}
        \toprule
        Dataset & \shortstack[c]{Latest \\ Model} & Scale & Generators & \shortstack[c]{High \\ Quality} & \shortstack[c]{Efficient \\ Encoding} & Resolution & FPS & Fairness & \shortstack[c]{Focus on \\ Real-World} & \shortstack[c]{In-the-wild \\ Data} \\
        \midrule
        GVD \cite{bai2024ai} & 2024/01 & 11k & 11 & \ding{55} & \ding{55} & Variable & - & - & - & \ding{51} \\
        GVF \cite{ma2024decof} & 2024/03 & 2.8k & 6 & \ding{55} & \ding{55} & Variable & - & - & - & \ding{55} \\
        GenVideo \cite{chen2024demamba} & 2024/04 & 2271k & 20 & \ding{55} & \ding{55} & Variable & 8-24 & - & - & \ding{51} \\
        DVF \cite{song2024learning} & 2024/10 & 7k & 8 & \ding{51} & \ding{55} & Variable & 8-30 & - & \ding{51} & \ding{55} \\
        \textbf{GenBuster} & \textbf{2026/02} & 203k & 23 & \ding{51} & \ding{51} & 1024x1024 & 24 & \ding{51} & \ding{51} & \ding{51} \\
        \bottomrule
    \end{tabular}
    }
    \label{tab:dataset_comparison}
    \vspace{-2em} 
\end{table}

To address these limitations, we introduce three contributions:

\noindent \textbf{1) GenBuster-200K: A High-Quality and Fair Dataset.} 
We introduce a curated dataset comprising over 200,000 high-resolution real and AI-generated video clips. Diverging from prior works, GenBuster-200K leverages the latest generators, enforces rigorous fairness constraints (e.g., balanced gender, ethnicity), and focuses heavily on high-fidelity, real-world scenarios. As shown in Table \ref{tab:dataset_comparison}, it successfully overcomes the authenticity and bias limitations of previous collections.

\noindent \textbf{2) GenBuster-Bench: A Progressive Diagnostic Benchmark.} 
To replace aggregated evaluations, we propose \textit{the first} benchmark that decomposes the video forgery detection challenge along two critical axes: \textit{domain shifts} and \textit{generational shifts}. This dual-axis design partitions the evaluation into three progressive tracks: \textit{In-Domain} (Era: 2024), \textit{Out-of-Domain} (Era: 2025), and \textit{In-the-Wild} (Era: 2026). Furthermore, we establish a standardized \textit{MLLM-as-a-Judge} protocol to exclusively evaluate the forensic grounding and reasoning depth of the MLLMs.

\noindent \textbf{3) BusterX: A Simple MLLM Baseline with RL Training.} 
We propose BusterX, which formulates forgery detection as a visual reasoning task where the reasoning chain itself serves as the detector. To the best of our knowledge, it is the \textit{first} MLLM-based video forgery detection framework with Reinforcement Learning (RL) training, empowering the model to uncover physical inconsistencies without expensive text annotations. Experimental results demonstrate that BusterX outperforms several leading MLLMs (e.g., Qwen3.5 \cite{qwen3.5}, Claude-Sonnet-4.6 \cite{claude-sonnet-4-6}) in both accuracy and interpretability.

\section{GenBuster-200K: A High-Quality and Fair Dataset}
\label{sec:dataset}

\textbf{Motivation.}
While recent pioneering datasets \cite{song2024learning, chen2024demamba} have catalyzed research in general video forgery detection, they are increasingly bottlenecked by outdated generative techniques, limited semantic diversity, and a heavy reliance on obviously fictional or low-realism content. To bridge the gap between current academic datasets and imminent real-world threats, we introduce GenBuster-200K. Designed to address these limitations, our dataset is constructed based on three principles: 1) \textbf{High Visual Quality}, as we employ state-of-the-art video generation models to ensure the synthetic videos exhibit high realism; 2) \textbf{Large Scale}, comprising over 200,000 high-resolution video clips; 3) \textbf{Real-World Focus and Fairness}. We deliberately filter out cartoonish prompts and enforce strict semantic constraints to ensure balanced demographic representations (e.g., gender, age, ethnicity), while actively mitigating harmful occupational and cultural stereotypes in practical, real-world scenarios.

\subsection{Data Construction}

\begin{figure}[t]
    \centering
    \includegraphics[width=1\linewidth]{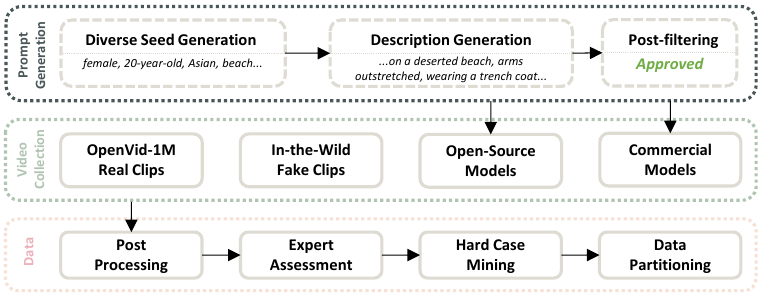}
    \caption{Data construction pipeline of GenBuster.}
    \label{fig:data_pipeline}
\end{figure}

\textbf{Data Sources.} GenBuster-200K consists of two parts: real videos and synthetic videos that simulate real-world conditions. Figure \ref{fig:data_pipeline} illustrates our data construction pipeline.

\noindent {\it Real Videos.} More than 100K real videos sourced from OpenVid-1M \cite{nan2024openvid}, covering a highly diverse spectrum of real-world, in-the-wild scenarios to ensure comprehensive coverage. These videos were carefully filtered to ensure high quality and eliminate any duplicate clips.

\noindent {\it Synthetic Videos.} More than 100K videos were generated with high realism and diversity. We adopt text-to-video generation, as it affords high semantic controllability. We developed a Synthetic Prompt Generation Pipeline (with full details provided in the appendix) and experimented with several generative models such as HunyuanVideo \cite{kong2024hunyuanvideo} and LTX-Video \cite{HaCohen2024LTXVideo}. For each model, we generated video frames with different hyper-parameters to further ensure visual diversity.

\noindent \textbf{Post-processing.} We perform unified post-processing \textbf{on both fake and real videos}. This process standardizes the resolution to 1024×1024, the duration to 5 seconds, and the frame rate to 24 FPS. This ensures consistency across videos generated by different models, which have varying original resolutions and frame rates. Additionally, we employ HEVC encoding using x265 with default parameters to encode the videos in the yuv420p10le format. This achieves a better balance between file size and video quality, allowing higher video quality at same bitrate. The unified post-processing eliminates potential biases from underlying encoding preferences (e.g., commonly used x264), which is crucial for ensuring the generalizability of forgery detection models across diverse sources. 
All videos are stored without audio after post-processing.

\noindent \textbf{Expert Assessment.} To ensure the fundamental quality and ethical alignment of our dataset, we engaged 5 well-trained annotators to perform rigorous quality control on the generated pool. Beyond technical filtering---such as removing corrupted video files, rendering failures, and clips with severe visual artifacts---this screening process included a strict \textbf{fairness audit}. Annotators were explicitly instructed to identify and discard videos exhibiting harmful demographic stereotypes, cultural misrepresentations, or unsafe content. Through this rigorous curation, we arrive at the final data distribution presented in Table \ref{tab:data_split}.

\subsection{Fairness-Aware Construction}
Adhering to the principle of \textbf{fairness}, our pipeline applies explicit semantic constraints during seed generation to promote demographic diversity in human-centric contexts. These constraints aim to balance gender representation, sample ages from a normal distribution, and include ethnic groups from diverse geographic regions. To further mitigate intersectional bias, we avoid disproportionately associating particular demographic groups with specific occupations, thereby reducing stereotypical demographic--occupational associations in the generated prompts.

\begin{table}[t]
    \centering
    \newcommand{\rotangle}{90}
    
    \caption{Data distribution for GenBuster.}
    \resizebox{\textwidth}{!}{
    \begin{tabular}{l cc cccc cccccccc ccccccccccc}
    \toprule
    \multirow{2}{*}[-3ex]{Split} & \multirow{2}{*}[-3ex]{Real} & \multirow{2}{*}[-3ex]{Fake} & 
    \multicolumn{4}{c}{ID} & 
    \multicolumn{8}{c}{OOD} & 
    \multicolumn{11}{c}{Wild} \\
    \cmidrule(lr){4-7} \cmidrule(lr){8-15} \cmidrule(lr){16-26}
    
    & & & 
    \rotatebox[origin=lb]{\rotangle}{CogVideoX} & \rotatebox[origin=lb]{\rotangle}{EasyAnimate} & \rotatebox[origin=lb]{\rotangle}{HunyuanVideo} & \rotatebox[origin=lb]{\rotangle}{LTX-Video} & 
    \rotatebox[origin=lb]{\rotangle}{Gen3} & \rotatebox[origin=lb]{\rotangle}{Jimeng} & \rotatebox[origin=lb]{\rotangle}{Kling} & \rotatebox[origin=lb]{\rotangle}{Luma} & \rotatebox[origin=lb]{\rotangle}{Pika} & \rotatebox[origin=lb]{\rotangle}{Sora} & \rotatebox[origin=lb]{\rotangle}{Vidu} & \rotatebox[origin=lb]{\rotangle}{Wanx} & 
    \rotatebox[origin=lb]{\rotangle}{Gen4.5} & \rotatebox[origin=lb]{\rotangle}{Hailuo2.3} & \rotatebox[origin=lb]{\rotangle}{Kling2.5 Turbo} & \rotatebox[origin=lb]{\rotangle}{Kling2.6} & \rotatebox[origin=lb]{\rotangle}{Pixverse v5} & \rotatebox[origin=lb]{\rotangle}{Seedance1.5 Pro} & \rotatebox[origin=lb]{\rotangle}{Seedance2.0} & \rotatebox[origin=lb]{\rotangle}{Sora2} & \rotatebox[origin=lb]{\rotangle}{Veo3.1} & \rotatebox[origin=lb]{\rotangle}{Wan2.5} & \rotatebox[origin=lb]{\rotangle}{Wan2.6} \\
    \midrule
    GenBuster-200K & 100k & 100k & 30k & 40k & 10k & 20k & - & - & - & - & - & - & - & - & - & - & - & - & - & - & - & - & - & - & - \\
    GenBuster-Bench: ID & 1000 & 1000 & 300 & 400 & 100 & 200 & - & - & - & - & - & - & - & - & - & - & - & - & - & - & - & - & - & - & - \\
    GenBuster-Bench: OOD & - & 1000 & - & - & - & - & 100 & 100 & 100 & 100 & 100 & 200 & 150 & 150 & - & - & - & - & - & - & - & - & - & - & - \\
    GenBuster-Bench: Wild & - & 150 & - & - & - & - & - & - & - & - & - & - & - & - & 10 & 20 & 15 & 20 & 10 & 20 & 15 & 10 & 10 & 10 & 10 \\
    \bottomrule
    \end{tabular}
    }
    \label{tab:data_split}
    \vspace{-1em}
\end{table}

\section{GenBuster-Bench: A Progressive Diagnostic Benchmark}
\label{sec:benchmark}

\subsection{Progressive Design}

Existing video forgery detection benchmarks \cite{ma2024decof} typically evaluate models on a monolithic test set. This aggregated approach is inherently flawed, as it obscures whether a model's failure stems from a lack of foundational forensic knowledge, poor cross-generator adaptability, or vulnerability to real-world degradations. Furthermore, generative techniques are \textbf{evolving} at an unprecedented pace, rendering static benchmarks quickly obsolete. To provide a comprehensive, diagnostic, and future-proof evaluation, we introduce GenBuster-Bench. It strategically decomposes the challenge along two critical axes---\textit{domain shifts} and \textit{generational shifts}---into three \textit{progressive} tracks:

\noindent {\it Level 1: In-Domain (ID) Track} -- Foundational Capability \textbf{(Era: 2024).} This track is directly constructed using the identical real video data sources and established open-source generators from GenBuster-200K. It serves as a standard test set to evaluate a detector's fundamental baseline ability to perceive known spatial-temporal artifacts on seen distributions and generators.

\noindent {\it Level 2: Out-of-Domain (OOD) Track} -- Generalization \textbf{(Era: 2025).} Attackers consistently leverage the newest, undisclosed generative models. To simulate this temporal arms race, the OOD track exclusively comprises videos from unseen, state-of-the-art commercial models representing the rapid capability leap of 2025. It strictly assesses whether a detector learns universal forensic laws rather than overfitting to outdated generator noise.

\noindent {\it Level 3: In-the-Wild (Wild) Track} -- Real-World Robustness \textbf{(Era: 2026).} Representing the most imminent and complex threat landscape, this is the ultimate stress test. We meticulously curated high-fidelity "hard negatives" from cutting-edge models and subjected them to dynamic real-world social media degradations. Furthermore, this track exclusively evaluates the \textit{explainability} of MLLMs strictly on correctly predicted samples, ensuring models can provide professional forensic rationales even when confronting the most deceptive future forgeries.

Through this dual-axis progressive paradigm, GenBuster-Bench transcends a simple leaderboard, serving as a robust diagnostic compass to evaluate the temporal lifespan and real-world reliability of future AI-generated video detectors.

\begin{figure}[t]
  \centering
  \includegraphics[width=0.97\linewidth]{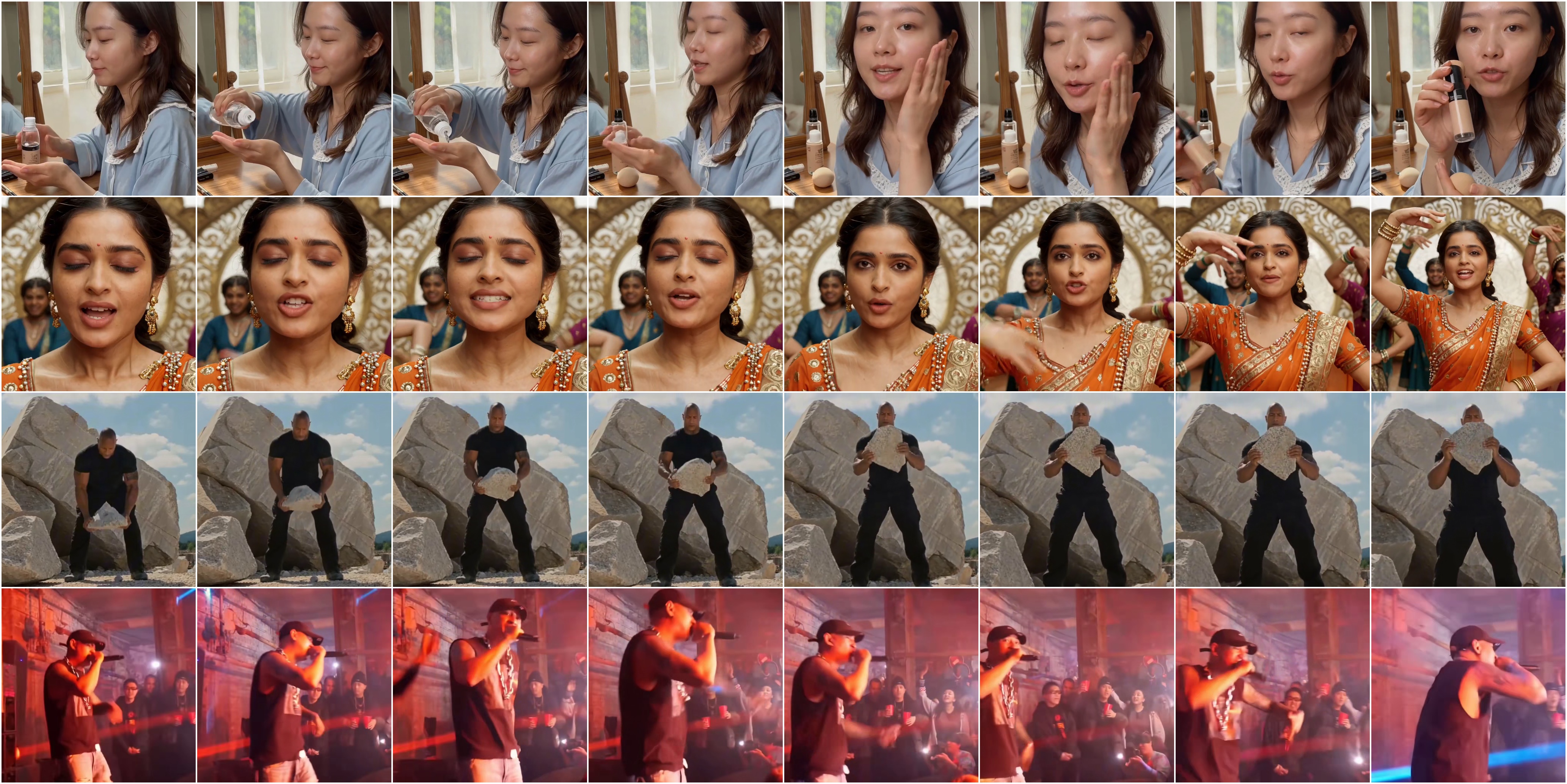}
  \caption{Visual Examples from GenBuster-Bench.}
  \label{fig:visual_examples}
\end{figure}

\subsection{Track Construction and Curation}

\noindent \textbf{Unified Pipeline.} To ensure a rigorous and standardized evaluation, all tracks in GenBuster-Bench strictly adopt the identical generation pipeline and post-processing protocols (both real and fake) as described for the GenBuster-200K. Figure \ref{fig:visual_examples} presents a selection of video examples from GenBuster, showcasing its high-quality and diversity.

\noindent \textbf{In-Domain (ID).} This track evaluates models on seen generators from GenBuster-200K. To prevent performance saturation and rigorously stress-test the detectors, the synthetic videos in this test set are purposefully generated at \textbf{higher specifications} (e.g., increased sampling steps) than training set.

\noindent \textbf{Out-of-Domain (OOD).} This track is designed to assess cross-generator generalization. It exclusively features videos from the unseen commercial models. Crucially, because many of these proprietary state-of-the-art models lack accessible APIs, our experts manually generated and collected these videos by directly interacting with their official web interfaces.

\noindent \textbf{In-the-Wild (Wild) Track.} Representing the most imminent threat landscape, the Wild track is specifically engineered to evaluate model robustness against real-world distributions. To construct this track, we collected high-fidelity forgeries actively circulating on prominent social platforms. Notably, this track incorporates content from the most advanced generative engines to date, such as Seedance 2.0. To maximize the forensic challenge, we implemented a highly stringent \textbf{expert-driven curation} process: from an initial pool of approximately 750 candidate videos generated by 2026-era models, our forensic experts manually selected only the top 150 samples. This selection process strictly eliminated videos with any perceptible artifacts, retaining only those that are visually indistinguishable from authentic footage to ensure a rigorous stress test.

\subsection{Evaluation Protocols and Metrics}

\noindent 
\textbf{Standardized Evaluation Protocol.} To ensure a fair and reproducible assessment across diverse architectures, we establish a standardized spatiotemporal sampling protocol. For all evaluated models, we sample frames at 2.0 FPS from each video. We generally resize video frames to a target resolution of approximately $256 \times 256$. However, recognizing the specific patch size constraints inherent to various Vision Transformers (ViTs), we allow the actual resized dimensions to vary slightly to strictly align with the closest multiple of each model's required patch size (e.g., $252 \times 252$ for Qwen2.5-VL).

\noindent
\textbf{Evaluation Metrics.} To comprehensively evaluate the performance of our model, we use overall accuracy (ACC) to measure the model's detection ability. Additionally, for every \textbf{unseen generator} in OOD track, we report its per-generator accuracy. Note that since our ID track is balanced between authentic and forged videos, we report the ACC on the \texttt{real} and \texttt{fake} subsets separately to expose potential model biases (e.g., over-forgery or conservative bias).

\section{BusterX: A Simple MLLM Baseline with RL Training}
\label{sec:method}

In this section, we introduce BusterX, an MLLM baseline for detecting AI-generated videos and providing detailed explanations using label-only, outcome-supervised RL. The training is general-purpose: it requires only video clips and binary real/fake labels, making it directly applicable to any forgery detection dataset without additional annotation.

\textbf{RL Training.} Benefiting from the powerful visual reasoning capabilities of modern MLLMs, we bypass the traditional supervised fine-tuning phase and \textbf{directly} proceed with RL training. We adopt DAPO \cite{yu2025dapoopensourcellmreinforcement} over GRPO \cite{deepseekai2025deepseekr1incentivizingreasoningcapability} primarily because DAPO effectively prevents \textit{entropy collapse} inherent to standard GRPO. Maintaining a healthy policy entropy is crucial, as it encourages the model to continuously explore the policy space and generate complex, multi-step reasoning chains instead of prematurely converging on superficial solutions.

\textbf{Reward Functions.} The total reward $R$ is a weighted sum of three components, designed to enforce format and encourage correct and detailed reasoning: $R = r_{\text{format}} + r_{\text{accuracy}} + r_{\text{soft\_overlong}}$

\begin{itemize}[leftmargin=*, nosep]
\item \textbf{Format}: If the model adheres to the specified format (i.e., \texttt{<think>...</think>\textbackslash n\textbackslash n...}), it receives a reward of $r_{\text{format}} = 0$; otherwise, it receives a penalty of $r_{\text{format}} = -1$.

\item \textbf{Accuracy}: If formatted and correct, it receives $r_{\text{accuracy}} = 1$; otherwise, $r_{\text{accuracy}} = 0$.

\item \textbf{Soft Overlong Penalty}: This component penalizes excessively long responses to prevent computational waste and mitigate repetitive token generation. The penalty is defined as:

\begin{equation}
    r_{\text{soft\_overlong}} =
    \begin{cases}
    0, & L_{\text{gen}} \le L_{\text{max}} - L_{\text{cache}} \\
    \frac{(L_{\text{max}} - L_{\text{cache}}) - L_{\text{gen}}}{L_{\text{cache}}}, & L_{\text{max}} - L_{\text{cache}} < L_{\text{gen}} \le L_{\text{max}} \\
    -1, & L_{\text{max}} < L_{\text{gen}}
    \end{cases}
    \label{eq:soft_punish}
\end{equation}

\end{itemize}

\section{Evaluation}
\label{sec:exp}

\begin{table}[t]
    \centering
    \caption{Comparison of state-of-the-art methods on GenBuster-Bench. The benchmark comprises three progressive tracks: ID evaluates performance on seen generators; OOD assesses cross-generator generalization on unseen commercial models; and Wild stress-tests real-world robustness. }
    \small
    \resizebox{1\textwidth}{!}{
    \begin{tabular}{l ccc cccccccccc cc}
    \toprule
        \multirow{2}{*}{\textbf{Methods}} & \multicolumn{3}{c}{\textbf{ID (2024)}} & \multicolumn{10}{c}{\textbf{OOD (2025)}} & \multicolumn{2}{c}{\textbf{Wild (2026)}} \\
        \cmidrule(lr){2-4} \cmidrule(lr){5-14} \cmidrule(l){15-16}
        & Real & Fake & ACC & Sora & Pika & Gen3 & Luma & WanX & Kling & Jimeng & Vidu & Fake & ACC & Fake & ACC \\
        \midrule
        
        \rowcolor{lightgray} \multicolumn{16}{l}{\textit{\textbf{Classical Detectors}}} \\
        3D ResNeXt \cite{hara3dcnns}
        & 84.4 & 60.8 & 72.6 & 31.0 & 75.0 & 39.0 & 42.0 & 48.7 & 22.0 & 65.0 & 46.0 & 44.7 & 64.6 & 39.3 & 61.9
        \\
        Vivit \cite{arnab2021vivit}
        & 92.9 & 64.1 & 78.5 & 74.0 & 52.0 & 69.0 & 53.0 & 62.0 & 55.0 & 33.0 & 68.7 & 60.6 & 76.8 & 46.7 & 69.8
        \\
        VideoMAE \cite{tong2022videomae}
        & 93.3 & 64.9 & 79.1 & 48.5 & 79.0 & 65.0 & 52.0 & 76.7 & 24.0 & 55.0 & 72.0 & 59.5 & 76.4 & 50.7 & 72.0
        \\
        DeMamba \cite{chen2024demamba}
        & \textbf{93.8} & 70.2 & 82.0 & 65.0 & 75.0 & 73.0 & 56.0 & 46.0 & 68.0 & 71.0 & 84.7 & 66.9 & 80.4 & 48.0 & 70.9
        \\
        
        \midrule
        \rowcolor{lightgray} \multicolumn{16}{l}{\textit{\textbf{Closed-Source MLLMs}}} \\
        GPT-5.2 (High) \cite{gpt5.2}
        & 93.3 & 61.8 & 77.6 & 43.0 & 77.0 & 78.0 & 91.0 & 66.7 & 36.0 & 27.0 & 33.3 & 54.5 & 73.9 & 50.7 & 72.0
        \\        
        Grok-4.1-fast \cite{grok4.1_fast}
        & 76.1 & 77.5 & 76.8 & 78.0 & 81.0 & 60.0 & 90.0 & 80.0 & 87.0 & \textbf{92.0} & 73.3 & 79.6 & 77.9 & 35.3 & 55.7
        \\
        Claude-Haiku-4.5 \cite{claude-haiku-4-5}
        & 84.7 & 58.7 & 71.7 & 53.0 & 40.0 & 72.0 & 80.0 & 60.0 & 31.0 & 46.0 & 46.7 & 53.5 & 69.1 & 19.3 & 52.0
        \\
        Claude-Sonnet-4.6 \cite{claude-sonnet-4-6}
        & 61.6 & 83.2 & 72.4 & 84.0 & \textbf{91.0} & \textbf{93.0} & 94.0 & 93.3 & 72.0 & 61.0 & 80.0 & 83.9 & 72.8 & 63.3 & 62.5
        \\

        \midrule 
        \rowcolor{lightgray} \multicolumn{16}{l}{\textit{\textbf{Open-Source MLLMs}}} \\
        Qwen3.5-4B \cite{qwen3.5}
        & 83.1 & 48.9 & 66.0 & 30.0 & 69.0 & 68.0 & 70.0 & 52.7 & 44.0 & 43.0 & 60.7 & 52.4 & 67.8 & 29.3 & 56.2
        \\
        Qwen3.5-27B \cite{qwen3.5}
        & 76.3 & 54.8 & 65.6 & 43.0 & 70.0 & 72.0 & 89.0 & 61.3 & 47.0 & 43.0 & 54.0 & 58.0 & 67.2 & 38.0 & 57.2
        \\
        Qwen3.5-122B-A10B \cite{qwen3.5}
        & 79.4 & 54.7 & 67.1 & 57.0 & 87.0 & 73.0 & 93.0 & 66.0 & 49.0 & 40.0 & 68.0 & 65.7 & 72.6 & 42.0 & 60.7
        \\
        
        \midrule
        \rowcolor{lightgray} \multicolumn{16}{l}{\textit{\textbf{domain-specific MLLMs}}} \\

        \textbf{BusterX}
        & 83.0 & \textbf{90.4} & \textbf{86.7} & \textbf{91.0} & 89.0 & 92.0 & \textbf{99.0} & \textbf{94.0} & \textbf{93.0} & 87.0 & \textbf{94.7} & \textbf{92.5} & \textbf{87.8} & \textbf{72.0} & \textbf{77.5}
        \\
        \bottomrule
    \end{tabular}
    }
    
    \raggedright
    \scriptsize{\textsuperscript{\dag} \textit{Note:} The ACC for the OOD and Wild tracks is calculated as the macro-average of the track's fake accuracy and the ID Real accuracy.}
    \label{tab:main_results}
    \vspace{-1em}
\end{table}

\textbf{Setup.} We adopt Qwen3.5-4B \cite{qwen3.5} as our base model and train BusterX on GenBuster-200K.
We employ the AdamW \citep{loshchilov2017decoupledadamw} with a learning rate of $1 \times 10^{-6}$, which aligns with established community best practices for stable RL training.
During training, we use DeepSpeed ZeRO, \texttt{bfloat16} precision.  
More details are shown in Appendix \ref{suppl:exp}.

\subsection{Main Results}
We evaluate SOTA baselines and BusterX across the three progressive tracks of GenBuster-Bench to assess detection accuracy and generalization. The detailed results are presented in Table~\ref{tab:main_results}. Before delving into specific track performances, we highlight our primary observation:

\begin{findingbox}
    \textit{\textbf{Finding:}} While traditional detectors and general MLLMs experience severe performance degradation on real-world manipulations (the \textit{Wild} track), BusterX achieves higher accuracy on this track than the evaluated baselines, supporting the effectiveness of its RL training approach.
\end{findingbox}

\noindent
\textbf{In-Domain (ID) Performance and Bias Mitigation.}
On the ID track, BusterX achieves the highest overall accuracy among all open-source MLLMs and classical detectors. More importantly, breaking down the ID performance into \texttt{Real} and \texttt{Fake} accuracy exposes severe inherent biases in existing models. For instance, Qwen3.5-122B exhibits a conservative bias (79.4\% on Real, but 54.7\% on Fake). Conversely, Claude-Sonnet-4.6 suffers from an over-forgery bias, which ruins its performance on real videos (61.6\% Real ACC). In contrast, BusterX maintains a balanced performance, demonstrating that our training paradigm successfully aligns the model's forensic perception without compromising its judgment on authentic videos.

\noindent
\textbf{Out-of-Domain (OOD) Generalization.}
The OOD track measures cross-generator generalization exclusively on unseen commercial models. BusterX achieves a remarkable 87.8\% accuracy, significantly outperforming all classical detectors and MLLMs.

\noindent
\textbf{In-the-Wild (Wild) Robustness.}
When confronting real-world social media in the Wild track, the performance of most classical detectors and generic MLLMs severely collapses (e.g., Claude-Haiku-4.5 drops to 19.3\% Fake). BusterX achieves 72.0\% Fake accuracy and 77.5\% ACC, outperforming the evaluated baselines on this track. These results support the effectiveness of BusterX under the challenging conditions represented by the Wild track. Notably, we also observe a clear scaling law within the Qwen3.5 family (4B, 27B, 122B): the detection accuracy on in-the-wild videos exhibits a steady upward trend as the model scales. This indicates that larger parameter sizes bring consistent benefits in handling highly diverse, real-world forgeries.

\vspace{1em}

\subsection{Explanation Evaluation}
\label{sec:explanation_eval}

\begin{wraptable}{r}{0.35\textwidth}
    \centering
    \caption{Explanation scores on correctly predicted Wild samples.}
    \small
    \resizebox{0.35\textwidth}{!}{
    \begin{tabular}{lc}
        \toprule
        \textbf{Model} & \textbf{Score} \\
        \midrule
        GPT-5.2 (High) \cite{gpt5.2} & 48.9 \\
        Grok-4.1-fast \cite{grok4.1_fast} & 81.0 \\
        Claude-Sonnet-4.6 \cite{claude-sonnet-4-6} & 78.3 \\
        \textbf{BusterX} & \textbf{84.7} \\
        \bottomrule
    \end{tabular}
    }
    \label{tab:explanation_results}
    \vspace{-1em}
\end{wraptable}

\textbf{MLLM-as-a-Judge.} 
We evaluate explanation quality using an MLLM-as-a-Judge experiment on the Wild track, employing Gemini 3 Flash Preview \cite{team2023gemini} as an impartial automated judge. To maintain strict visual consistency, the judge evaluates the rationales using the exact same frame sequences provided to the models during the detection phase. Crucially, evaluating the rationale of an incorrect prediction is inherently flawed, as it forces the model to hallucinate justifications for a false premise. To prevent this, our protocol is \textit{strictly accuracy-decoupled}—we exclusively prompt the judge to evaluate explanations for samples from the Wild track where the evaluated model successfully predicted the ground-truth label. The judge scores the generated rationales on a discrete 1-5 scale across three critical dimensions: 1) \textbf{Forensic Visual Grounding}, which assesses the identification of systemic rendering artifacts rather than superficial entity descriptions; 2) \textbf{Reasoning Depth \& Logic Structure}, which evaluates cognitive depth and causal physical reasoning; and 3) \textbf{Factual Accuracy \& Pseudo-Precision Penalty}, which penalizes hallucinations and unverifiable claims (e.g., falsely citing specific frame numbers). The final explainability score is derived by averaging these dimensions and normalizing the result to the $[0, 1]$ interval.
The full judge prompt is provided in Appendix~\ref{suppl:judge}.
Table~\ref{tab:explanation_results} compares explanation scores for four representative MLLMs. BusterX achieves the highest score of 84.7 among these MLLMs.

\noindent
\textbf{False-Positive Analysis.}
Across all false-positive rationales, 89.6\% explicitly contain terms such as ``render,'' ``3D,'' or ``simulation,'' suggesting a tendency to attribute visual effects and stylized content to AI generation. We manually reviewed 20 false-positive cases by comparing the cited evidence with the video frames: 16 contained observable cues incorrectly attributed to AI generation, while 4 contained hallucinated evidence absent from the video. Representative errors include treating a floating visual-effect icon or a 2D explosion effect as evidence of AI generation, and misinterpreting hands and motion textures in a natural scene as generation artifacts. These cases distinguish incorrect attribution of visible evidence from hallucination of the evidence itself.

\subsection{Case Study}
\begin{figure}[!t]
   \centering
   \includegraphics[width=1\linewidth]{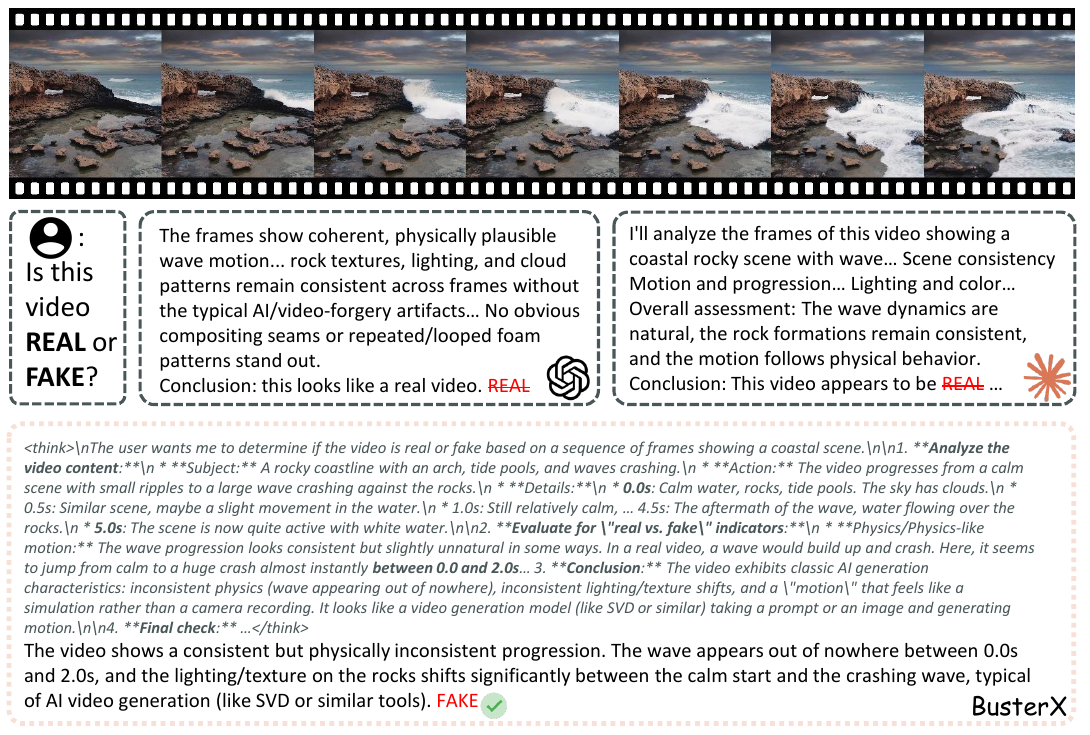}
   \caption{Case Study. Faced with highly realistic synthetic video, MLLMs like GPT-5.2 and Claude-Sonnet-4.6 fail, whereas BusterX successfully performs frame-level spatiotemporal localization to uncover underlying physical inconsistencies. Full responses are provided in Appendix \ref{suppl:complete}.}
   \label{fig:case_study}
\end{figure}

As illustrated in Figure~\ref{fig:case_study}, we evaluate BusterX against advanced proprietary MLLMs (e.g., GPT-5.2 and Claude-Sonnet-4.6) when faced with highly realistic synthetic videos. While commercial models often rely on superficial visual coherence and fail to detect deep fakes, BusterX demonstrates superior frame-level spatiotemporal localization and deep reasoning grounded in world knowledge. Our qualitative analysis reveals three key capabilities:

\textbf{1) Uncovering Physical Inconsistencies via Spatiotemporal Grounding:} In the coastal wave video (Figure~\ref{fig:case_study}), baseline MLLMs incorrectly assess the video as real, fooled by seemingly coherent rock textures and wave motion. Conversely, BusterX isolates fatal spatiotemporal anomalies. It explicitly tracks the timeline, noting that the water transitions from calm to a massive crashing wave abruptly between 0.0s and 2.0s. Furthermore, BusterX identifies unnatural shifts in lighting and texture.

\textbf{2) Assessing Logical Consistency:} BusterX's ability to track long-horizon logic is highly effective in complex urban scenes (Appendix~\ref{suppl:additional}). When analyzing a synthetic street video featuring a tram, BusterX catches severe object permanence violations---specifically, a crowd of pedestrians spontaneously appearing on a previously empty platform at the 5.0s mark.

\textbf{3) Validating Realism through World Knowledge:} When analyzing real-world footage (Appendix~\ref{suppl:additional}), BusterX does not just look for low-level artifacts; it actively interprets the specific event content by grounding its visual observations in world knowledge. For instance, in an outdoor cooking scenario, the model identifies specific cultural and environmental contexts, such as the subject's traditional attire (kurta) and the rural setting framed by a dirt ground and rice paddies. Furthermore, it applies culinary common sense to evaluate the scene's logic, and even questioning whether the observed cooking techniques align with real-world practices for making fritters (vadas). By corroborating these semantic details with practical physical indicators, BusterX comprehensively verifies both the event logic and physical realism to confidently authenticate genuine videos.

\begin{figure}[htbp]
    \centering
    \begin{minipage}[t]{0.32\textwidth}
        \centering
        \captionof{table}{Robustness evaluation under various degradations.}
        \small
        \resizebox{1\linewidth}{!}{
        \begin{tabular}{lccc}
        \toprule
            \textbf{Degradation} & \textbf{ID} & \textbf{OOD} & \textbf{Wild} \\
            \midrule
            \textbf{None} & \textbf{86.7} & \textbf{87.8} & \textbf{77.5} \\ 
            \midrule
            50\% Frame    & 84.8 & 86.2 & 76.0 \\
            25\% Frame    & 83.7 & 85.5 & 76.2 \\
            JPEG 80       & 86.0 & 87.8 & 76.8 \\
            JPEG 70       & 85.4 & 87.3 & 77.1 \\
            \bottomrule
        \end{tabular}
        }
        \label{tab:robustness}
    \end{minipage}\hfill
    \begin{minipage}[t]{0.32\textwidth}
        \centering
        \captionof{table}{Ablation on \\ w/o Thinking and RL loss type.}
        \small
        \resizebox{1\linewidth}{!}{
        \begin{tabular}{lccc}
        \toprule
            \textbf{Variant} & \textbf{ID} & \textbf{OOD} & \textbf{Wild} \\
            \midrule
            \textbf{BusterX} & \textbf{86.7} & \textbf{87.8} & \textbf{77.5} \\ 
            \midrule
            w/o Thinking   & 69.4 & 79.9 & 70.0 \\
            w/ GRPO loss   & 85.9 & 85.1 & 70.9 \\
            \bottomrule
        \end{tabular}
        }
        \label{tab:ablation}
    \end{minipage}\hfill
    \begin{minipage}[t]{0.35\textwidth}
        \centering
        \captionof{figure}{Ablation on reward funcs.}
        \vspace{-0.5em}
        \includegraphics[width=0.9\linewidth]{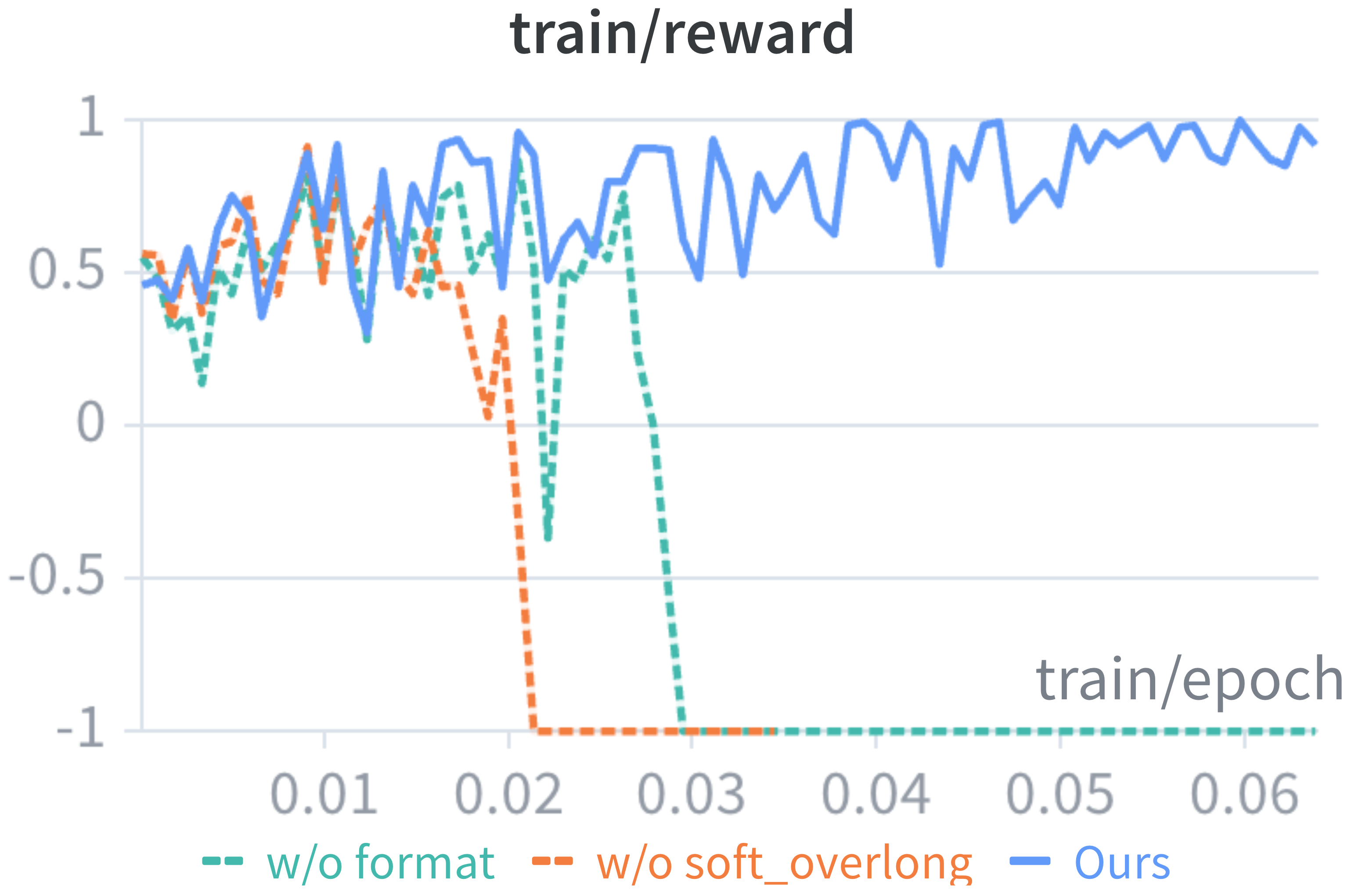}
        \label{fig:reward_func_ablation}
    \end{minipage}
    \vspace{-1em}
\end{figure}

\subsection{Ablation and Robustness Studies}

\textbf{Reasoning chain matters.} As detailed in Table \ref{tab:ablation}, skipping the reasoning chain (w/o Thinking) causes a severe performance collapse across all tracks. This confirms our core premise that formulating detection as a step-by-step visual reasoning task is essential.


\textbf{DAPO vs. GRPO.} Table \ref{tab:ablation} also highlights the impact of the RL objective. Replacing DAPO with standard GRPO loss noticeably degrades accuracy. We observe that this drop is primarily driven by severe \textit{entropy collapse} during GRPO training. This premature collapse drastically restricts the policy space, severely hindering the model's ability to explore and formulate the in-depth physical reasoning chains crucial for forgery detection. By effectively mitigating this collapse, DAPO ensures robust exploration and optimization of the reasoning process. More details are provided in Appendix \ref{suppl:dapo}.

\textbf{Ablation on Reward functions.} As shown in Figure \ref{fig:reward_func_ablation}, we evaluate the training stability across different reward configurations. Without the soft overlong penalty (w/o soft\_overlong) or the format constraint (w/o format), the training exhibits significant fragility; it fails to converge and often collapses into repetitive patterns, demonstrating the necessity of our reward design.

\textbf{Robustness.} Table~\ref{tab:robustness} demonstrates BusterX maintains consistent performance against video perturbations without specific degradation training, ensuring real-world practicality.

\section{Related Work}
\label{sec:related_work}

\subsection{AI-Generated Video Detection Datasets}
Early datasets primarily focused on face manipulation and generation, typically employing GANs, VAEs, or various swapping techniques to produce synthetic videos. Datasets such as FaceForensics++ \cite{rossler2019faceforensicspp}, FakeAVCeleb \cite{khalid2021fakeavceleb}, WildDeepfake \cite{zi2020wilddeepfake}, and KoDF \cite{Kwon_2021_ICCV} are categorized as deepfake video detection dataset \cite{ni2025genvidbench}. In recent years, identity-driven forgery datasets have also emerged, with IDForge \cite{xu2024identity} being a representative example.
Recent advances in generative technology have significantly improved the quality of video generation, expanding synthetic content beyond facial manipulations to encompass full-scene natural video synthesis. This shift has broadened the research focus from facial forgeries to holistic video generation, driving the development of diverse datasets for AI-generated video detection. For instance, GenVideo \cite{chen2024demamba} and GenVidBench \cite{ni2025genvidbench} are large-scale benchmarks built on multiple AI generation models, covering a diverse range of semantic content. However, they are often limited by issues such as relatively low visual quality and lack of fairness. In contrast, GenBuster addresses these challenges by offering higher-quality data, a larger volume, emphasis on fairness, and focus on real-world applicability.

\subsection{AI-Generated Video Detection Methods}
Previous methods \cite{wang2020cnn,ojha2023towards,wang2023dynamic,duan2024test,tan2024rethinking,tan2023learning} primarily formulated AI-generated content detection as a binary classification problem, focusing mainly on image-based deepfake detection. These methods utilized various architectures, such as CNNs and Transformers. To enhance the interpretability of the models, Huang et al. \cite{Huang2024SIDASM} recently propose a method for localizing altered regions in images and generating textual explanations, with a specific emphasis on detecting image editing manipulations.
As AI-generated video quality advances, the demand for robust detection methods grows increasingly urgent. Early works primarily focused on face-centric deepfake detection \cite{zheng2021exploring,concas2024quality,pellicer2024pudd, DBLP:journals/corr/abs-2410-06126}. For instance, Zheng et al.~\cite{zheng2021exploring,concas2024quality} develop methods that identify spatial inconsistencies to distinguish real from synthetic content. Meanwhile, Pellicer et al.~\cite{pellicer2024pudd} propose a model that learns person-specific prototypes by analyzing patterns in real videos. 
However, these methods struggle to generalize to in-the-wild AI-generated video detection. To address these limitations, recent works have explored alternative strategies. DeMamba \cite{chen2024demamba} detects synthetic videos by analyzing small spatiotemporal zones, capturing subtle pixel changes. UNITE \cite{kundu2024towards} enhances synthetic video detection by capturing full-frame manipulations using a transformer-based architecture with domain-agnostic features. Reiss et al. \cite{reiss2023detecting} propose the concept of fact-checking for detection and introduce FACTOR, a practical recipe that effectively counters critical attacks such as face swapping and audio-visual synthesis.
To address these challenges, BusterX not only detects video authenticity but also provides detailed explanations by treating AI-generated video detection as a visual reasoning task, thereby enhancing the detection analysis.

\subsection{Multimodal Large Language Models}
MLLMs have evolved significantly in recent years. For instance, BLIP-2 \cite{li2023blip} and LLaVA \cite{liu2023visual} integrated vision encoders with large autoregressive language models, enabling image-conditioned text generation and multimodal reasoning. Qwen2.5-VL \cite{bai2025qwen2} further advances these developments by improving vision-language alignment through a more efficient vision encoder and enhanced cross-modal fusion mechanisms. Concurrently, there is growing interest in leveraging MLLMs for fine-grained pixel-level understanding tasks \cite{zhang2025pixelsailsingletransformerpixelgrounded, yuan2025sa2vamarryingsam2llava, ren2024pixellm, ren2025videoworld, ren2026videoworld2}.
Recently, the integration of MLLMs has introduced a new paradigm in AIGC detection \cite{zou2025surveyaigeneratedmediadetection}. For image forgery detection, models such as AntifakePrompt \cite{DBLP:journals/corr/abs-2310-17419}, ForgeryGPT \cite{liu2025forgerygptmultimodallargelanguage}, FakeShield \cite{xu2024fakeshieldexplainableimageforgery}, ForgeryTalker \cite{lian2024largescaleinterpretablemultimodalitybenchmark}, ForgerySleuth \cite{sun2024forgerysleuthempoweringmultimodallarge}, and SIDA \cite{Huang2024SIDASM} utilize the capabilities of MLLMs to enhance detection accuracy and provide robust support for localization and explanation tasks. In the video domain, MM-Det \cite{song2024learning} leverages the strong visual-linguistic representations of MLLMs by generating a Multi-Modal Forgery Representation, which enhances its ability to detect unseen manipulated content.
Reinforcement Learning (RL) has demonstrated its effectiveness in enhancing LLM reasoning capabilities \cite{achiam2023gpt,team2023gemini} following the Supervised Fine-Tuning (SFT) stage \cite{luo2023wizardmath}. Notably, DeepSeek-R1 \cite{deepseekai2025deepseekr1incentivizingreasoningcapability} introduced Group Relative Policy Optimization (GRPO), a variant of the Proximal Policy Optimization (PPO) algorithm \cite{schulman2017proximal}, to significantly improve model reasoning \cite{DBLP:journals/corr/abs-2402-03300}. The community has made numerous recent efforts to further refine GRPO, with DAPO \cite{yu2025dapoopensourcellmreinforcement} and GSPO \cite{zheng2025gspo} being representative examples. To enable the detection model to not only classify videos as real or fake but also provide explanations that align with human understanding, we apply RL fine-tuning to the video forgery detection task, representing the first attempt of this paradigm in the research domain.

\section{Conclusion}
\label{sec:Conclusion}

In this paper, we address critical vulnerabilities in AI-generated video detection through three contributions: the GenBuster-200K dataset, the GenBuster-Bench benchmark, and the BusterX detection framework. GenBuster-200K addresses the authenticity and distributional limitations found in prior datasets. GenBuster-Bench provides structured diagnostic evaluation along temporal and domain axes, and introduces an MLLM-as-a-judge to assess forensic explainability independently of classification performance. BusterX leverages reinforcement learning to train MLLMs toward transparent, step-by-step physical reasoning, moving beyond opaque classification outputs.

\noindent \textbf{Limitations.} Generated explanations may hallucinate details or misattribute observable cues to AI generation. Verifying these explanations against video frames assesses visual grounding rather than causal faithfulness. Our framework is also limited to the visual modality, leaving audio manipulation and audiovisual semantic inconsistencies unaddressed — both of which are common in real-world deepfake content. Future work will extend toward audio-visual multimodal detection.

\section*{Acknowledgments}
This work was funded by the Engineering and Physical Sciences Research Council [grant number EP/X525741/1] Impact Acceleration Accounts Grant.

{
    \small
    \bibliographystyle{plainnat}
    \bibliography{reference}

@String(CVPR= {IEEE Conf. Comput. Vis. Pattern Recog.})

@String(ICCV= {Int. Conf. Comput. Vis.})

@String(ACMMM= {ACM Int. Conf. Multimedia})

@String(ICLR = {Int. Conf. Learn. Represent.})

@String(CVPR  = {CVPR})

@String(ICCV  = {ICCV})

@String(ACMMM = {ACM MM})

@String(ICLR  = {ICLR})

@String{ICML = {ICML}}

@string{ARXIV = {Arxiv}}

@article{DBLP:journals/corr/abs-2402-03300,
  author       = {Zhihong Shao and
                  Peiyi Wang and
                  Qihao Zhu and
                  Runxin Xu and
                  Junxiao Song and
                  Mingchuan Zhang and
                  Y. K. Li and
                  Y. Wu and
                  Daya Guo},
  title        = {DeepSeekMath: Pushing the Limits of Mathematical Reasoning in Open
                  Language Models},
  journal      = {Arxiv},
  year         = {2024},
}

@inproceedings{rossler2019faceforensicspp,
  title={Faceforensics++: Learning to detect manipulated facial images},
  author={Rossler, Andreas and Cozzolino, Davide and Verdoliva, Luisa and Riess, Christian and Thies, Justus and Nie{\ss}ner, Matthias},
  booktitle={ICCV},
  pages={1--11},
  year={2019}
}

@article{DBLP:journals/corr/abs-2410-06126,
    author       = {Yize Chen and Zhiyuan Yan and Siwei Lyu and Baoyuan Wu},
    title        = {\emph{X}\({}^{\mbox{2}}\)-DFD: {A} framework for e\emph{X}plainable and e\emph{X}tendable Deepfake Detection},
    journal={Arxiv},
    year         = {2024},
}

@article{DBLP:journals/tmm/JuJCGL24,
  author       = {Yan Ju and
                  Shan Jia and
                  Jialing Cai and
                  Haiying Guan and
                  Siwei Lyu},
  title        = {{GLFF:} Global and Local Feature Fusion for AI-Synthesized Image Detection},
  journal      = {{IEEE} Trans. Multim.},
  year         = {2024},
}

@inproceedings{wang2020cnn,
  title={CNN-generated images are surprisingly easy to spot... for now},
  author={Wang, Sheng-Yu and Wang, Oliver and Zhang, Richard and Owens, Andrew and Efros, Alexei A},
  booktitle={CVPR},
  year={2020}
}

@inproceedings{ojha2023towards,
  title={Towards universal fake image detectors that generalize across generative models},
  author={Ojha, Utkarsh and Li, Yuheng and Lee, Yong Jae},
  booktitle={CVPR},
  year={2023}
}

@inproceedings{tan2024rethinking,
  title={Rethinking the up-sampling operations in cnn-based generative network for generalizable deepfake detection},
  author={Tan, Chuangchuang and Zhao, Yao and Wei, Shikui and Gu, Guanghua and Liu, Ping and Wei, Yunchao},
  booktitle={CVPR},
  year={2024}
}

@inproceedings{tan2023learning,
  title={Learning on gradients: Generalized artifacts representation for gan-generated images detection},
  author={Tan, Chuangchuang and Zhao, Yao and Wei, Shikui and Gu, Guanghua and Wei, Yunchao},
  booktitle={CVPR},
  year={2023}
}

@inproceedings{Huang2024SIDASM,
  title={SIDA: Social Media Image Deepfake Detection, Localization and Explanation with Large Multimodal Model}, 
  author={Zhenglin Huang and Jinwei Hu and Xiangtai Li and Yiwei He and Xingyu Zhao and Bei Peng and Baoyuan Wu and Xiaowei Huang and Guangliang Cheng},
  booktitle={CVPR},
  year={2025}
}

@inproceedings{song2024learning,
  title={On learning multi-modal forgery representation for diffusion generated video detection},
  author={Song, Xiufeng and Guo, Xiao and Zhang, Jiache and Li, Qirui and Bai, Lei and Liu, Xiaoming and Zhai, Guangtao and Liu, Xiaohong},
  booktitle={NeurIPS},
  year={2024}
}

@inproceedings{wang2023dynamic,
  title={Dynamic graph learning with content-guided spatial-frequency relation reasoning for deepfake detection},
  author={Wang, Yuan and Yu, Kun and Chen, Chen and Hu, Xiyuan and Peng, Silong},
  booktitle={CVPR},
  year={2023}
}

@article{duan2024test,
  title={Test-time Forgery Detection with Spatial-Frequency Prompt Learning},
  author={Duan, Junxian and Ai, Yuang and Liu, Jipeng and Huang, Shenyuan and Huang, Huaibo and Cao, Jie and He, Ran},
  journal={ICCV},
  year={2024},
}

@article{kundu2024towards,
  title={Towards a Universal Synthetic Video Detector: From Face or Background Manipulations to Fully AI-Generated Content},
  author={Kundu, Rohit and Xiong, Hao and Mohanty, Vishal and Balachandran, Athula and Roy-Chowdhury, Amit K},
  journal={Arxiv},
  year={2024},
}

@article{reiss2023detecting,
  title={Detecting deepfakes without seeing any},
  author={Reiss, Tal and Cavia, Bar and Hoshen, Yedid},
  journal={Arxiv},
  year={2023},
}

@inproceedings{concas2024quality,
  title={Quality-based artifact modeling for facial deepfake detection in videos},
  author={Concas, Sara and La Cava, Simone Maurizio and Casula, Roberto and Orru, Giulia and Puglisi, Giovanni and Marcialis, Gian Luca},
  booktitle={CVPR},
  year={2024}
}

@inproceedings{pellicer2024pudd,
  title={PUDD: towards robust multi-modal prototype-based deepfake detection},
  author={Pellicer, Alvaro Lopez and Li, Yi and Angelov, Plamen},
  booktitle={CVPR},
  year={2024}
}

@inproceedings{zheng2021exploring,
  title={Exploring temporal coherence for more general video face forgery detection},
  author={Zheng, Yinglin and Bao, Jianmin and Chen, Dong and Zeng, Ming and Wen, Fang},
  booktitle={CVPR},
  year={2021}
}

@article{achiam2023gpt,
  title={Gpt-4 technical report},
  author={Achiam, Josh and Adler, Steven and Agarwal, Sandhini and Ahmad, Lama and Akkaya, Ilge and Aleman, Florencia Leoni and Almeida, Diogo and Altenschmidt, Janko and Altman, Sam and Anadkat, Shyamal and others},
  journal={Arxiv},
  year={2023}
}

@article{team2023gemini,
  title={Gemini: a family of highly capable multimodal models},
  author={Team, Gemini and Anil, Rohan and Borgeaud, Sebastian and Alayrac, Jean-Baptiste and Yu, Jiahui and Soricut, Radu and Schalkwyk, Johan and Dai, Andrew M and Hauth, Anja and Millican, Katie and others},
  journal={Arxiv},
  year={2023}
}

@inproceedings{li2023blip,
  title={Blip-2: Bootstrapping language-image pre-training with frozen image encoders and large language models},
  author={Li, Junnan and Li, Dongxu and Savarese, Silvio and Hoi, Steven},
  booktitle={ICML},
  year={2023},
}

@inproceedings{liu2023visual,
  title={Visual instruction tuning},
  author={Liu, Haotian and Li, Chunyuan and Wu, Qingyang and Lee, Yong Jae},
  booktitle={NeurIPS},
  year={2023}
}

@article{bai2025qwen2,
  title={Qwen2. 5-VL Technical Report},
  author={Bai, Shuai and Chen, Keqin and Liu, Xuejing and Wang, Jialin and Ge, Wenbin and Song, Sibo and Dang, Kai and Wang, Peng and Wang, Shijie and Tang, Jun and others},
  journal={Arxiv},
  year={2025}
}

@article{deepseekai2024deepseekv3technicalreport,
      title={DeepSeek-V3 Technical Report}, 
      author={DeepSeek-AI},
      year={2024},
      journal={Arxiv},
}

@article{deepseekai2025deepseekr1incentivizingreasoningcapability,
      title={DeepSeek-R1: Incentivizing Reasoning Capability in LLMs via Reinforcement Learning}, 
      author={DeepSeek-AI},
      year={2025},
      journal={Arxiv},
}

@article{yu2025dapoopensourcellmreinforcement,
      title={DAPO: An Open-Source LLM Reinforcement Learning System at Scale}, 
      author={Qiying Yu and Zheng Zhang and Ruofei Zhu and Yufeng Yuan and Xiaochen Zuo and Yu Yue and Tiantian Fan and Gaohong Liu and Lingjun Liu and Xin Liu and Haibin Lin and Zhiqi Lin and Bole Ma and Guangming Sheng and Yuxuan Tong and Chi Zhang and Mofan Zhang and Wang Zhang and Hang Zhu and Jinhua Zhu and Jiaze Chen and Jiangjie Chen and Chengyi Wang and Hongli Yu and Weinan Dai and Yuxuan Song and Xiangpeng Wei and Hao Zhou and Jingjing Liu and Wei-Ying Ma and Ya-Qin Zhang and Lin Yan and Mu Qiao and Yonghui Wu and Mingxuan Wang},
      year={2025},
      journal={Arxiv},
}

@article{touvron2023llamaopenefficientfoundation,
      title={LLaMA: Open and Efficient Foundation Language Models}, 
      author={Hugo Touvron and Thibaut Lavril and Gautier Izacard and Xavier Martinet and Marie-Anne Lachaux and Timothée Lacroix and Baptiste Rozière and Naman Goyal and Eric Hambro and Faisal Azhar and Aurelien Rodriguez and Armand Joulin and Edouard Grave and Guillaume Lample},
      year={2023},
      journal={Arxiv},
}

@article{khalid2021fakeavceleb,
  title={{FakeAVCeleb}: A Novel Audio-Video Multimodal Deepfake Dataset}, 
  author={Hasam Khalid and Shahroz Tariq and Minha Kim and Simon S. Woo},
  year={2022},
  journal={Arxiv},
}

@inproceedings{zi2020wilddeepfake,
  title={Wilddeepfake: A challenging real-world dataset for deepfake detection},
  author={Zi, Bojia and Chang, Minghao and Chen, Jingjing and Ma, Xingjun and Jiang, Yu-Gang},
  booktitle={ACMMM},
  year={2020}
}

@InProceedings{Kwon_2021_ICCV,
    author    = {Kwon, Patrick and You, Jaeseong and Nam, Gyuhyeon and Park, Sungwoo and Chae, Gyeongsu},
    title     = {KoDF: A Large-Scale Korean DeepFake Detection Dataset},
    booktitle = {ICCV},
    year      = {2021},
}

@article{chen2024demamba,
  title={Demamba: Ai-generated video detection on million-scale genvideo benchmark},
  author={Chen, Haoxing and Hong, Yan and Huang, Zizheng and Xu, Zhuoer and Gu, Zhangxuan and Li, Yaohui and Lan, Jun and Zhu, Huijia and Zhang, Jianfu and Wang, Weiqiang and others},
  journal={Arxiv},
  year={2024}
}

@inproceedings{hara3dcnns,
  author={Kensho Hara and Hirokatsu Kataoka and Yutaka Satoh},
  title={Can Spatiotemporal 3D CNNs Retrace the History of 2D CNNs and ImageNet?},
  booktitle={CVPR},
  year={2018},
}

@inproceedings{xu2024identity,
  title={Identity-Driven Multimedia Forgery Detection via Reference Assistance},
  author={Xu, Junhao and Chen, Jingjing and Song, Xue and Han, Feng and Shan, Haijun and Jiang, Yu-Gang},
  booktitle={ACMMM},
  year={2024}
}

@article{ni2025genvidbench,
  title={GenVidBench: A Challenging Benchmark for Detecting AI-Generated Video},
  author={Ni, Zhenliang and Yan, Qiangyu and Huang, Mouxiao and Yuan, Tianning and Tang, Yehui and Hu, Hailin and Chen, Xinghao and Wang, Yunhe},
  journal={Arxiv},
  year={2025}
}

@article{ma2024decof,
  title={Detecting AI-Generated Video via Frame Consistency},
  author={Long Ma and Zhiyuan Yan and Qinglang Guo and Yong Liao and Haiyang Yu and Pengyuan Zhou},
  journal=ARXIV,
  year={2024}
}

@article{nan2024openvid,
  title={Openvid-1m: A large-scale high-quality dataset for text-to-video generation},
  author={Nan, Kepan and Xie, Rui and Zhou, Penghao and Fan, Tiehan and Yang, Zhenheng and Chen, Zhijie and Li, Xiang and Yang, Jian and Tai, Ying},
  journal={Arxiv},
  year={2024}
}

@inproceedings{bai2024ai,
  title={AI-Generated Video Detection via Spatial-Temporal Anomaly Learning},
  author={Bai, Jianfa and Lin, Man and Cao, Gang and Lou, Zijie},
  booktitle={PRCV},
  year={2024},
}

@article{kong2024hunyuanvideo,
      title={HunyuanVideo: A Systematic Framework For Large Video Generative Models}, 
      author={Kong, Weijie and Tian, Qi and Zhang, Zijian and Min, Rox and Dai, Zuozhuo and Zhou, Jin and Xiong, Jiangfeng and Li, Xin and Wu, Bo and Zhang, Jianwei and Wu, Kathrina and Lin, Qin and Wang, Aladdin and Wang, Andong and Li, Changlin and Huang, Duojun and Yang, Fang and Tan, Hao and Wang, Hongmei and Song, Jacob and Bai, Jiawang and Wu, Jianbing and Xue, Jinbao and Wang, Joey and Yuan, Junkun and Wang, Kai and Liu, Mengyang and Li, Pengyu and Li, Shuai and Wang, Weiyan and Yu, Wenqing and Deng, Xinchi and Li, Yang and Long, Yanxin and Chen, Yi and Cui, Yutao and Peng, Yuanbo and Yu, Zhentao and He, Zhiyu and Xu, Zhiyong and Zhou, Zixiang and Xu, Zunnan and Tao, Yangyu and Lu, Qinglin and Liu, Songtao and Zhou, Dax and Wang, Hongfa and Yang, Yong and Wang, Di and Liu, Yuhong and Jiang, Jie and Zhong, Caesar},
      year={2024},
      journal={Arxiv},
}

@article{HaCohen2024LTXVideo,
  title={LTX-Video: Realtime Video Latent Diffusion},
  author={HaCohen, Yoav and Chiprut, Nisan and Brazowski, Benny and Shalem, Daniel and Moshe, Dudu and Richardson, Eitan and Levin, Eran and Shiran, Guy and Zabari, Nir and Gordon, Ori and Panet, Poriya and Weissbuch, Sapir and Kulikov, Victor and Bitterman, Yaki and Melumian, Zeev and Bibi, Ofir},
  journal={Arxiv},
  year={2024}
}

@misc{Luma,
  title = {Luma},
  howpublished = {\url{https://lumalabs.ai}}
}

@inproceedings{arnab2021vivit,
  title={Vivit: A video vision transformer},
  author={Arnab, Anurag and Dehghani, Mostafa and Heigold, Georg and Sun, Chen and Lu{\v{c}}i{\'c}, Mario and Schmid, Cordelia},
  booktitle={CVPR},
  year={2021}
}

@inproceedings{tong2022videomae,
  title={Videomae: Masked autoencoders are data-efficient learners for self-supervised video pre-training},
  author={Tong, Zhan and Song, Yibing and Wang, Jue and Wang, Limin},
  booktitle={NeurIPS},
  year={2022}
}

@article{luo2023wizardmath,
  title={Wizardmath: Empowering mathematical reasoning for large language models via reinforced evol-instruct},
  author={Luo, Haipeng and Sun, Qingfeng and Xu, Can and Zhao, Pu and Lou, Jianguang and Tao, Chongyang and Geng, Xiubo and Lin, Qingwei and Chen, Shifeng and Zhang, Dongmei},
  journal={Arxiv},
  year={2023}
}

@article{schulman2017proximal,
  title={Proximal policy optimization algorithms},
  author={Schulman, John and Wolski, Filip and Dhariwal, Prafulla and Radford, Alec and Klimov, Oleg},
  journal={Arxiv},
  year={2017}
}

@article{yuan2025sa2vamarryingsam2llava,
      title={Sa2VA: Marrying SAM2 with LLaVA for Dense Grounded Understanding of Images and Videos}, 
      author={Haobo Yuan and Xiangtai Li and Tao Zhang and Zilong Huang and Shilin Xu and Shunping Ji and Yunhai Tong and Lu Qi and Jiashi Feng and Ming-Hsuan Yang},
      year={2025},
      journal={Arxiv},
}

@article{zhang2025pixelsailsingletransformerpixelgrounded,
      title={Pixel-SAIL: Single Transformer For Pixel-Grounded Understanding}, 
      author={Tao Zhang and Xiangtai Li and Zilong Huang and Yanwei Li and Weixian Lei and Xueqing Deng and Shihao Chen and Shunping Ji and Jiashi Feng},
      year={2025},
      journal={Arxiv},
}

@article{DBLP:journals/corr/abs-2310-17419,
  author       = {You{-}Ming Chang and
                  Chen Yeh and
                  Wei{-}Chen Chiu and
                  Ning Yu},
  title        = {AntifakePrompt: Prompt-Tuned Vision-Language Models are Fake Image
                  Detectors},
  journal      = {Arxiv},
  year         = {2023}
}

@article{xu2024fakeshieldexplainableimageforgery,
      title={FakeShield: Explainable Image Forgery Detection and Localization via Multi-modal Large Language Models}, 
      author={Zhipei Xu and Xuanyu Zhang and Runyi Li and Zecheng Tang and Qing Huang and Jian Zhang},
      year={2024},
      journal={Arxiv},
}

@article{liu2025forgerygptmultimodallargelanguage,
    title={ForgeryGPT: Multimodal Large Language Model For Explainable Image Forgery Detection and Localization}, 
    author={Jiawei Liu and Fanrui Zhang and Jiaying Zhu and Esther Sun and Qiang Zhang and Zheng-Jun Zha},
    year={2024},
    journal={Arxiv},
}

@article{lian2024largescaleinterpretablemultimodalitybenchmark,
      title={A Large-scale Interpretable Multi-modality Benchmark for Facial Image Forgery Localization}, 
      author={Jingchun Lian and Lingyu Liu and Yaxiong Wang and Yujiao Wu and Li Zhu and Zhedong Zheng},
      year={2024},
      journal={Arxiv},
}

@article{sun2024forgerysleuthempoweringmultimodallarge,
      title={ForgerySleuth: Empowering Multimodal Large Language Models for Image Manipulation Detection}, 
      author={Zhihao Sun and Haoran Jiang and Haoran Chen and Yixin Cao and Xipeng Qiu and Zuxuan Wu and Yu-Gang Jiang},
      year={2024},
      journal={Arxiv},
}

@article{zou2025surveyaigeneratedmediadetection,
      title={Survey on AI-Generated Media Detection: From Non-MLLM to MLLM}, 
      author={Yueying Zou and Peipei Li and Zekun Li and Huaibo Huang and Xing Cui and Xuannan Liu and Chenghanyu Zhang and Ran He},
      year={2025},
      journal={Arxiv},
}

@inproceedings{kwon2023vllm,
  title={Efficient Memory Management for Large Language Model Serving with PagedAttention},
  author={Woosuk Kwon and Zhuohan Li and Siyuan Zhuang and Ying Sheng and Lianmin Zheng and Cody Hao Yu and Joseph E. Gonzalez and Hao Zhang and Ion Stoica},
  booktitle={ACM SIGOPS},
  year={2023}
}

@article{zheng2025gspo,
  title={Group sequence policy optimization},
  author={Zheng, Chujie and Liu, Shixuan and Li, Mingze and Chen, Xiong-Hui and Yu, Bowen and Gao, Chang and Dang, Kai and Liu, Yuqiong and Men, Rui and Yang, An and others},
  journal=ARXIV,
  year={2025}
}

@misc{seedance2.0,
  author = {{ByteDance}},
  title = {Seedance 2.0},
  year = {2026},
  url = {https://seed.bytedance.com/en/seedance2_0}
}

@misc{qwen3.5,
    title  = {{Qwen3.5}: Towards Native Multimodal Agents},
    author = {{Qwen Team}},
    year   = {2026},
    url    = {https://qwen.ai/blog?id=qwen3.5}
}

@misc{gpt5.2,
    title  = {Introducing GPT‑5.2},
    author = {OpenAI},
    year   = {2025},
    url    = {https://openai.com/index/introducing-gpt-5-2/}
}

@misc{grok4.1_fast,
    title  = {Grok 4.1 Fast and Agent Tools API},
    author = {X.AI},
    year   = {2025},
    url    = {https://x.ai/news/grok-4-1-fast/}
}

@misc{claude-haiku-4-5,
    title  = {Introducing Claude Haiku 4.5},
    author = {Anthropic},
    year   = {2026},
    url    = {https://www.anthropic.com/news/claude-haiku-4-5/}
}

@misc{claude-sonnet-4-6,
    title  = {Introducing Claude Sonnet 4.6},
    author = {Anthropic},
    year   = {2026},
    url    = {https://www.anthropic.com/news/claude-sonnet-4-6/}
}

@inproceedings{loshchilov2017decoupledadamw,
title={Decoupled weight decay regularization},
author={Loshchilov, Ilya and Hutter, Frank},
booktitle=ICLR,
year={2019}
}

@inproceedings{ren2026videoworld2,
  title={Videoworld 2: Learning transferable knowledge from real-world videos},
  author={Ren, Zhongwei and Wei, Yunchao and Yu, Xiao and Luo, Guixun and Zhao, Yao and Kang, Bingyi and Feng, Jiashi and Jin, Xiaojie},
  booktitle=CVPR,
  year={2026}
}

@inproceedings{ren2025videoworld,
  title={Videoworld: Exploring knowledge learning from unlabeled videos},
  author={Ren, Zhongwei and Wei, Yunchao and Guo, Xun and Zhao, Yao and Kang, Bingyi and Feng, Jiashi and Jin, Xiaojie},
  booktitle=CVPR,
  year={2025}
}

@inproceedings{ren2024pixellm,
  title={Pixellm: Pixel reasoning with large multimodal model},
  author={Ren, Zhongwei and Huang, Zhicheng and Wei, Yunchao and Zhao, Yao and Fu, Dongmei and Feng, Jiashi and Jin, Xiaojie},
  booktitle=CVPR,
  year={2024}
}
}

\appendix
\section{Societal Impact}
\label{suppl:societal_impact}

The proliferation of AI-generated video poses tangible risks to public discourse, including the spread of misinformation, non-consensual synthetic media, and manipulation of political or legal evidence. This work contributes tools intended to help counter these harms. At the same time, we acknowledge the dual-use nature of this research. Detailed forensic analysis of generator artifacts may inadvertently inform efforts to produce harder-to-detect content. We release our benchmark and model with the expectation that they will be used to strengthen detection capabilities, and encourage the community to consider access controls when deploying forensic tools in adversarial settings.

\section{Prompt Generation Pipeline in GenBuster-200K}
\label{suppl:prompt_gen_pipeline} 

The main process comprises three stages:

\noindent {\it Stage-1: Diverse Seed Generation.} This stage is explicitly engineered for \textbf{Fairness}: keyword collection begins with human-centric contexts where we systematically enforce balanced sampling across gender, age, ethnicity and intersecting attributes, then extends to animal, scenery and other non-human domains to ensure the resulting semantic space is globally inclusive and bias-resistant.
This methodical approach ensures our initial semantic space is both inclusive and comprehensive, spanning a wide range of perspectives.

\noindent {\it Stage-2: Description Generation.} Building on the prior output from stage-1, the forthcoming synthetic video descriptions will be generated by an LLM using a predefined prompt. Specifically, the process includes:
\textbf{1)} Prompt Constraints Guidance: The prompt introduces physical and aesthetic constraints, with manual adjustments to cover multiple scenarios. These constraints guide the LLM to generate descriptions that are both realistic and contextually appropriate.
\textbf{2)} Multi-model Overfitting Resistance: To combat model-specific bias, we randomly sample from a diverse pool of over 20 open-source LLMs, spanning varied architectures and parameter scales (e.g., DeepSeek-V3 \cite{deepseekai2024deepseekv3technicalreport}, LLaMA \cite{touvron2023llamaopenefficientfoundation}, etc.). We employ high-temperature sampling to maximize output diversity during generation.

\noindent {\it Stage-3: Post-filtering.} After generating the prompt, we employ a multi-step post-filtering mechanism to ensure the quality and reliability of the generated prompts. Specifically, the post-filtering process includes the following key steps:
\textbf{1)} Length Control: We first check the length of each generated prompt to ensure it falls within a predefined range. Prompts that are too short may lack sufficient context, while those that are excessively long can introduce unnecessary complexity or redundancy. 
\textbf{2)} Special Character Check: We then scan for any special characters or formatting issues that may have been inadvertently introduced during the generation process. 
\textbf{3)} Readability Scoring: The LLM Judge assigns a readability score based on linguistic coherence, grammatical correctness, and overall clarity. This step is crucial because low-quality prompts (those with poor readability) tend to produce low-quality generated videos that are easily detectable, which would undermine the dataset's utility for training robust detectors. More details are provided in the appendix.
By integrating these steps, we can ensure that the generated prompts are of high quality.

\section{DAPO}
\label{suppl:dapo}

We utilize DAPO \cite{yu2025dapoopensourcellmreinforcement} as our reinforcement learning strategy. DAPO samples a group of outputs $\{o_i\}_{i=1}^G$ for each question $q$ from the old policy $\pi_{\theta_{old}}$. The reward model is then used to score the outputs, yielding $\{r_i\}_{i=1}^G$ correspondingly. Then DAPO optimizes the policy model $\pi_{\theta}$ by maximizing the following objective:

\begin{equation}
\begin{aligned}
\mathcal{J}_{\text{DAPO}}(\theta) =\quad& \mathbb{E}_{(q,a)\sim \mathcal{D}, \{o_i\}_{i=1}^G\sim \pi_{\theta_\text{old}}(\cdot\mid q)}\\&
\Bigg[\frac{1}{\sum_{i=1}^{G}|o_i|}\sum_{i=1}^{G}\sum_{t=1}^{|o_i|} 
\min \Big( r_{i,t}(\theta) \hat{A}_{i,t},\\&
\ \text{clip} \Big( r_{i,t}(\theta), 1 - {\varepsilon_{\text{low}}}, 1 + {\varepsilon_{\text{high}}} \Big) \hat{A}_{i,t} \Big) \Bigg]
\\
\text{s.t.}\quad& 0< \Big|\{o_i\mid\texttt{is\_equivalent}(a,o_i)\}\Big|< G,
\label{eq:dapoloss}
\end{aligned}
\end{equation}
where $\varepsilon_{\text{low}}$ and $\varepsilon_{\text{high}}$ are hyper-parameters, and $\hat{A}_{i,t}$ is the advantage:
\begin{equation}
    r_{i,t}(\theta)=\frac{\pi_{\theta}(o_{i,t} \mid q, o_{i,<t})}{\pi_{\theta_{\text{old}}}(o_{i,t} \mid q,o_{i,<t})},
    \quad\hat{A}_{i,t} = \frac{r_i - \text{mean}(\{r_i\}_{i=1}^G)}{\text{std}(\{r_i\}_{i=1}^G)}.
\label{eq:advantage_calculation}
\end{equation}

\begin{figure}[h]
  \centering
  \includegraphics[width=0.65\linewidth]{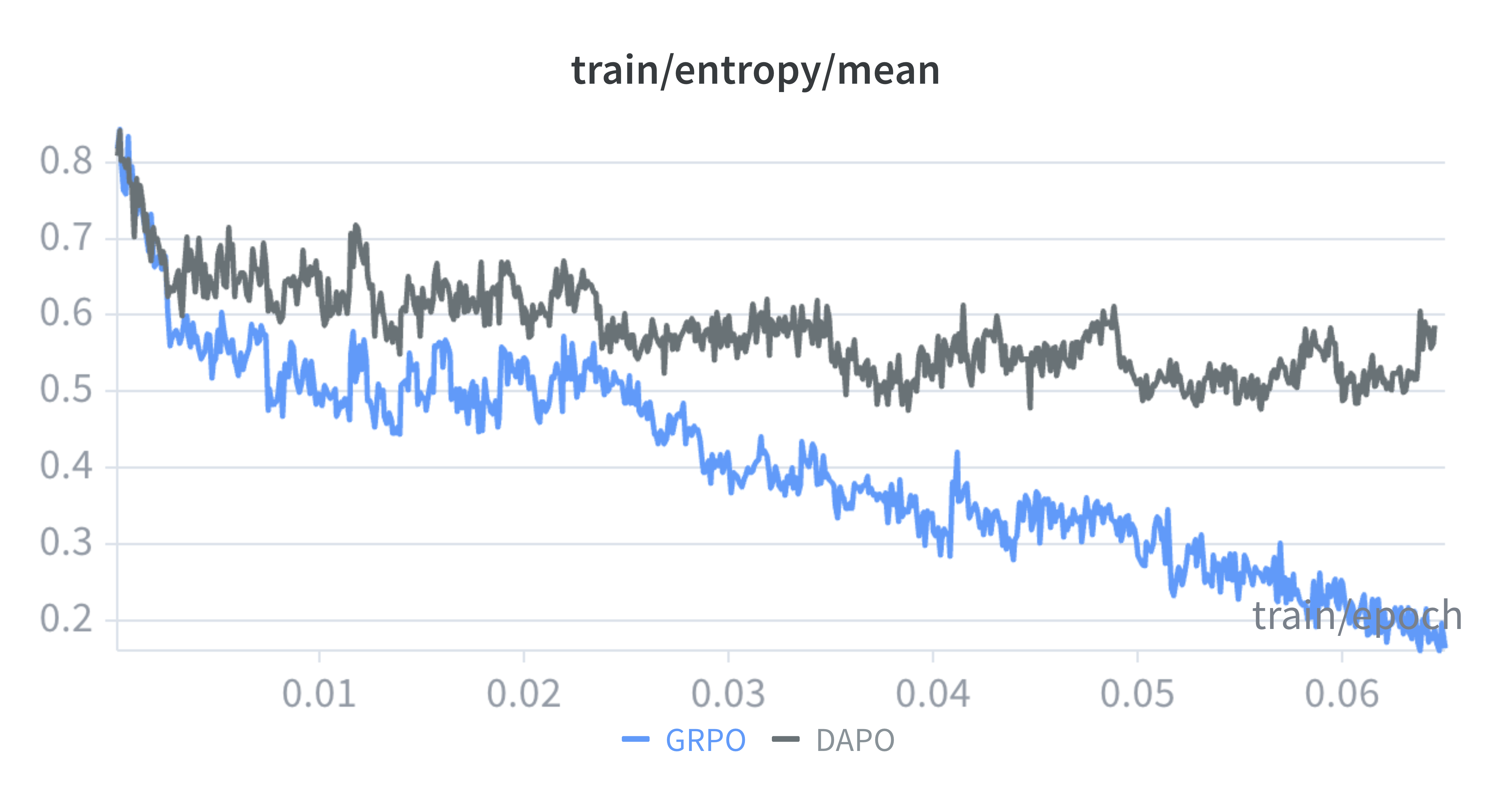}
  \caption{Training entropy: DAPO vs. GRPO.}
  \label{fig:grpo_entropy_collapse}
\end{figure}

As shown in Figure \ref{fig:grpo_entropy_collapse}, DAPO effectively prevents entropy collapse inherent to standard GRPO.

\section{Experimental Details}
\label{suppl:exp}

In this section, we provide comprehensive details regarding the training configurations, evaluation parameters, and hyperparameter selections utilized in our study to ensure full reproducibility.

To minimize generation overhead, we integrate \texttt{vLLM} \citep{kwon2023vllm} for both training rollouts and evaluation. Its high-throughput capabilities enable us to maintain a large training rollout batch size. For each query, we generate $G=8$ trajectories at a temperature of $T=1.0$. The maximum completion length is set to $4,096$ tokens. We apply a clipping bound of $[0.8, 1.28]$ for the policy ratio.

\section{Prompt Design}
\label{suppl:prompt}

Based on extensive experiments, we have carefully designed the following prompt to guide the answer.

\begin{tcolorbox}[title = {User Prompt}, fontupper=\small]
{
\it 
Please analyze whether there are any inconsistencies or obvious signs of forgery in the video, and finally come to a conclusion: Is this video real or fake?\\

Then, just answer this MCQ with a single letter:\\
Q: Is this video real or fake?\\
Options:\\
A) real\\
B) fake\\

Put the final answer in $\backslash \backslash boxed\{...\}$
}
\end{tcolorbox}

\newpage

\section{MLLM-as-a-judge}
\label{suppl:judge}

To rigorously evaluate the rationale and forensic reasoning capabilities of the models, we employ \textbf{Gemini 3 Flash Preview} as Judge. The full prompt template is detailed below:

\begin{tcolorbox}[title = {User Prompt}, fontupper=\tiny]
{
\it 
Please carefully watch the provided video. \\
Here is the rationale generated by the evaluated model: \\

"\{explanation\}"\\

Evaluate the rationale strictly using the following 1-5 discrete rubric for three dimensions.\\

---\\
\#\#\# Dimension 1: Forensic Visual Grounding (Systemic vs. Superficial)\\
Evaluates whether the model merely describes objects (image captioning) or genuinely grounds its analysis in systemic forensic artifacts (lighting, physics, rendering).\\
* 1 (Vague/Captioning): Mentions nothing specific, or just describes the scene's content (e.g., "there is a man and a car") without identifying actual forensic flaws.\\
* 2 (Coarse Entity): Mentions specific entities being "fake" or "unnatural" without explaining the physical manifestation.\\
* 3 (Local Feature): Locks onto isolated local features (e.g., "blurry face").\\
* 4 (Holistic/Systemic Grounding): Perfectly identifies systemic rendering artifacts rather than just objects. This includes artificial lighting uniformity, lack of real-world analog camera noise/entropy, overly perfect textures, or static motion rendering.\\
* 5 (Advanced Spatiotemporal): Seamlessly combines spatial artifacts with temporal dynamics, or provides a master-level grounding of cinematic, environmental, and textural inconsistencies across the sequence.\\

\#\#\# Dimension 2: Reasoning Depth \& Logic Structure\\
Evaluates the cognitive depth, use of deductive reasoning, and structural rigor of the analysis.\\
* 1 (Basic/Tautological): "It is fake because it is AI-generated" or zero logical flow.\\
* 2 (Surface Flaw): Points out a flaw but provides no causal or physical reasoning.\\
* 3 (Mechanism-Level): Explains the flaw using basic generative defect mechanisms.\\
* 4 (Deductive Physical Analysis): Contrasts the observed digital rendering against expected real-world physics (e.g., "Real-world videos would have natural shadows... but this shows uniform artificial lighting").\\
* 5 (Structured Forensic Chain-of-Thought): Employs a highly structured, step-by-step reasoning process (e.g., breaking down lighting, posture, background, and camera artifacts systematically). It reads like a methodical, professional digital forensic investigation report.\\

\#\#\# Dimension 3: Factual Accuracy \& Pseudo-Precision Penalty\\
Evaluates if the model fabricated details or used deceptive "pseudo-precision" to appear accurate. Compare strictly against the ACTUAL video frames.\\
* 1 (Severe Hallucination / Pseudo-Precision): The model fabricates key evidence. **CRITICAL PENALTY:** If the model acts like a "frame-counter" and cites highly specific frame numbers (e.g., "In frame 5...", "In frame 11...") or instantaneous color swaps that you CANNOT strictly verify in your limited visual sample, it is hallucinating precision and MUST receive a 1.\\
* 2 (Partial Over-interpretation): Half factual, half forced correlation.\\
* 3 (Slight Hallucination): Mostly correct, but slightly mischaracterizes a minor visual detail.\\
* 4 (Mostly Accurate): Perfectly aligns with the video content without inventing unverifiable data points.\\
* 5 (Absolutely Rigorous \& Objective): Fully grounded in the true pixels, lighting, and overarching physics of the video. Zero hallucination, avoiding unverifiable pseudo-precision, maintaining an objective forensic tone.\\

---\\
\#\#\# Output JSON Format\\
You must return the evaluation strictly in the following JSON format. Provide a brief analysis (1-2 sentences) before assigning the integer score (1-5) for each dimension.\\

\{\\
  "dim1": \{\\
    "analysis": "<your step-by-step reasoning>",\\
    "score": <1-5>\\
  \},\\
  "dim2": \{\\
    "analysis": "<your step-by-step reasoning>",\\
    "score": <1-5>\\
  \},\\
  "dim3": \{\\
    "analysis": "<your step-by-step reasoning>",\\
    "score": <1-5>\\
  \}\\
\}\\
}
\end{tcolorbox}

\newpage

\section{Complete Responses}
\label{suppl:complete}

The model's full responses in Case Study. This is a fake video, the answer should be $B$.

\begin{figure}[h!]
  \centering
  \includegraphics[width=1.0\linewidth]{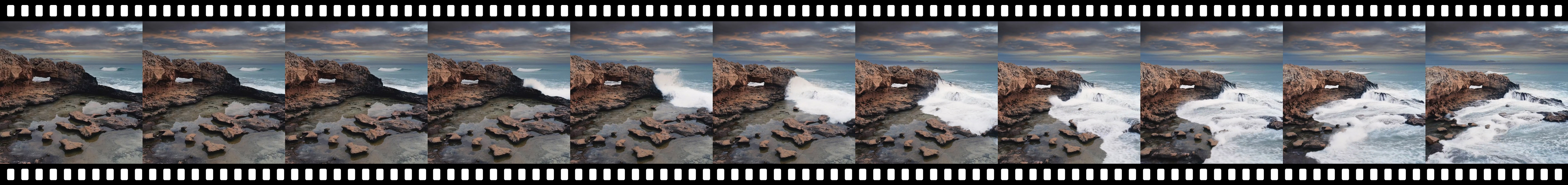}
  \vspace{-2em}
  \label{fig:hard_case_film_strip}
\end{figure}

\begin{tcolorbox}[title = {GPT-5.2}, fontupper=\small]
{
The frames show coherent, physically plausible wave motion (cresting, impact against the rock face, spray/foam expansion, then foam dispersal and flowing around the tide pool rocks). The rock textures, lighting, and cloud patterns remain consistent across frames without the typical AI/video-forgery artifacts (warping edges, temporal “melting,” inconsistent reflections, or unstable fine details). No obvious compositing seams or repeated/looped foam patterns stand out.\\

Conclusion: this looks like a real video.\\

$\backslash \backslash boxed\{A\}$
}
\end{tcolorbox}

\begin{tcolorbox}[title = {Claude-Sonnet-4.6}, fontupper=\small]
{
I'll analyze the frames of this video showing a coastal rocky scene with waves.\\

\#\# Analysis\\

**Scene consistency:**
- The location shows a rocky coastal area with a natural rock arch, tidal pools, and ocean waves
- The rock formations, arch structure, and general geography remain consistent throughout all frames\\

**Motion and progression:**
- The video shows a natural wave sequence: calm water → approaching wave → wave crashing against rocks → water spreading/receding
- The wave movement follows physically plausible fluid dynamics
- The timing and progression of the wave appears natural\\

**Lighting and color:**
- There is a gradual, subtle shift in lighting/color temperature across frames (slightly cooler to slightly warmer tones)
- This could indicate time-lapse or color grading, but appears consistent with natural lighting changes at dusk/dawn\\

**Potential anomalies:**
- The color grading shifts somewhat dramatically between early and later frames
- However, this appears to be natural lighting variation during the golden hour/dusk period
- Water movement, foam patterns, and spray all appear physically consistent
- No obvious digital artifacts, cloning, or unnatural elements detected\\

**Overall assessment:**
The wave dynamics are natural, the rock formations remain consistent, and the motion follows expected physical behavior. The color shifts appear to be natural lighting changes during twilight conditions.\\

\#\# Conclusion
This video appears to be **real** - a genuine timelapse/video of ocean waves crashing against a rocky coastal formation with a natural arch.\\

$\backslash \backslash boxed\{A\}$
}
\end{tcolorbox}

\newpage

\begin{tcolorbox}[title = {BusterX}, fontupper=\tiny]
{
<think>\\The user wants me to determine if the video is real or fake based on a sequence of frames showing a coastal scene.\\\\1.  **Analyze the video content:**\\    *   **Subject:** A rocky coastline with an arch, tide pools, and waves crashing.\\    *   **Action:** The video progresses from a calm scene with small ripples to a large wave crashing against the rocks.\\    *   **Details:**\\        *   0.0s: Calm water, rocks, tide pools. The sky has clouds.\\        *   0.5s: Similar scene, maybe a slight movement in the water.\\        *   1.0s: Still relatively calm, but the water looks a bit more turbulent.\\        *   1.5s: A wave is starting to form in the distance.\\        *   2.0s: A wave is breaking near the shore.\\        *   2.5s: The wave is crashing, creating white foam.\\        *   3.0s: The wave is fully crashing against the rocks.\\        *   3.5s: The wave is receding slightly but still turbulent.\\        *   4.0s: More white foam, the wave has hit the rocks.\\        *   4.5s: The aftermath of the wave, water flowing over the rocks.\\        *   5.0s: The scene is now quite active with white water.\\\\2.  **Evaluate for real vs. fake indicators:**\\    *   **Physics/Physics-like motion:** The wave progression looks consistent but slightly unnatural in some ways. In a real video, a wave would build up and crash. Here, it seems to jump from calm to a huge crash almost instantly between 0.0 and 2.0s. In reality, a wave takes time to build. The transition from calm water (0.0) to a crashing wave (2.5) happens very quickly, suggesting a generated video or a very specific edit.\\    *   **Texture/Details:** Look at the rocks and the water.\\        *   The rocks look consistent but have a slightly smooth, almost painted or rendered texture in some frames (especially 0.0-1.0). In later frames, they look more like realistic rocks. Wait, looking closely at the rocks, they seem consistent but the lighting changes drastically.\\        *   The water texture: In 0.0-1.0, it's calm and reflective. In 2.0+, it's churning white foam. This is a common AI generation technique (motion generation).\\        *   **The most telling part:** The morph or transition. The video starts with a still, almost like a photo, and then suddenly a wave crashes. In a real video, you'd see the wave building up. Here, it's almost like a video generation prompt where the prompt is a rocky coast with a wave crashing. The lighting changes significantly between 0.0 and 5.0. In 0.0, it's a soft, maybe slightly overcast or sunset light. In 5.0, the lighting seems similar but the motion blur and foam suggest a different pass.\\    *   **Specific AI artifacts:**\\        *   Look at the texturing of the rocks. In 0.0, the rock on the far left is distinct. In 5.0, it's in the same spot but the texture looks slightly different or just different lighting.\\        *   Look at the tide pools or shallow water. In 0.0, you see reflections. In 5.0, the water is churning.\\        *   **The most obvious fake sign:** The motion of the wave. In 0.0-1.0, the water is almost glassy and calm. In 2.0, a white blob appears. By 2.5, it's a full crash. This is a classic motion or video generation artifact where the model tries to simulate physics but often gets the timing wrong. A real wave wouldn't just pop into existence like that. It would build up.\\        *   **The sky:** The clouds move, but the lighting on the rocks changes significantly between the calm start and the crash. In 0.0, the sun seems low (sunset/sunrise). In 5.0, it's still low but the lighting on the wet rocks changes drastically, suggesting a different render or a very inconsistent motion simulation.\\\\3.  **Conclusion:** The video exhibits classic AI generation characteristics: inconsistent physics (wave appearing out of nowhere), inconsistent lighting/texture shifts, and a motion that feels like a simulation rather than a camera recording. It looks like a video generation model (like SVD or similar) taking a prompt or an image and generating motion.\\\\4.  **Final check:** The prompt asks for a single letter MCQ answer.\\    *   Q: Is this video real or fake?\\    *   Options: A) real B) fake\\    *   Answer: B\\</think>\\\\The video shows a consistent but physically inconsistent progression. The wave appears out of nowhere between 0.0s and 2.0s, and the lighting/texture on the rocks shifts significantly between the calm start and the crashing wave, typical of AI video generation (like SVD or similar tools).\\

$\backslash \backslash boxed\{B\}$
}
\end{tcolorbox}

\newpage

\section{Additional Examples of BusterX}
\label{suppl:additional}

\begin{figure}[h!]
  \centering
  \includegraphics[width=1.0\linewidth]{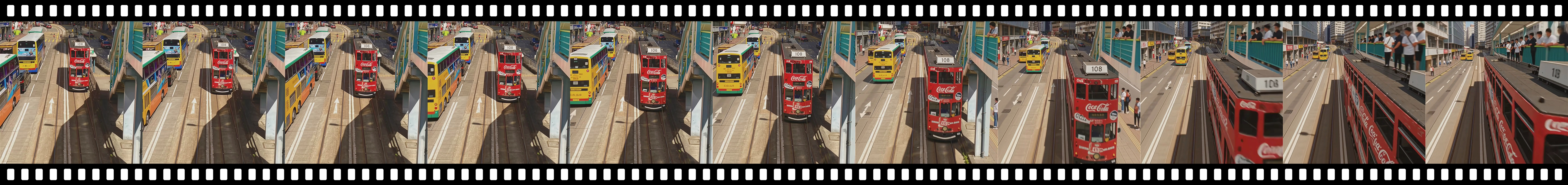}
  \vspace{-2em}
  \label{fig:fake_case_film_strip}
\end{figure}

\begin{tcolorbox}[title = {FAKE}, fontupper=\tiny]
{
<think>\\The user wants me to determine if the video is real or fake based on a sequence of frames showing a street scene with buses and a tram.\\\\1.  **Analyze the video content:**\\    *   **Subject:** A red Coca-Cola tram is moving along tracks. There are other buses (yellow, blue) on the road next to it. People are walking on the sidewalk and waiting at a platform.\\    *   **Motion:** The tram is moving forward. The buses are moving or stopped. People are walking.\\    *   **Details:**\\        *   0.0s: Tram is moving forward. A yellow bus is behind it. A blue bus is further back. People are on the sidewalk.\\        *   0.5s: Similar scene. The tram is moving. The yellow bus is moving.\\        *   1.0s: The tram is moving. The yellow bus is moving.\\        *   1.5s: The tram is moving. The yellow bus is moving.\\        *   2.0s: The tram is moving. The yellow bus is moving.\\        *   2.5s: The tram is moving. The yellow bus is moving.\\        *   3.0s: The tram is moving. The yellow bus is moving. People are on the platform.\\        *   3.5s: The tram is moving. The yellow bus is moving. People are on the platform.\\        *   4.0s: The tram is moving. The yellow bus is moving. People are on the platform.\\        *   4.5s: The tram is moving. The yellow bus is moving. People are on the platform.\\        *   5.0s: The tram is moving. The yellow bus is moving. People are on the platform.\\\\2.  **Evaluate for real vs. fake indicators:**\\    *   **Physics/Motion:** The motion looks consistent but slightly unnatural in some frames. For example, in 0.0s, the tram is moving forward. In 5.0s, it's still moving forward. The cars and buses are moving in a way that suggests a simulation. Look at the wheels and reflections.\\    *   **Object details:**\\        *   The tram has Coca-Cola branding clearly visible. It's a specific model (looks like a double-decker tram).\\        *   The buses are also distinct. One is yellow and blue, another is orange and blue.\\        *   The environment looks like a city street (Hong Kong maybe? The architecture and the tram suggest it).\\    *   **Key AI/Generative artifacts:**\\        *   **Textures:** Look at the road surface and the tram tracks. In 0.0s, the tracks are clear. In 5.0s, the tracks are clear. But look at the reflections on the tram in 0.0 vs 5.0. In 0.0, it's reflecting the surroundings. In 5.0, it's reflecting the surroundings.\\        *   **Motion blur/Physics:** Look at the people. In 0.0, people are walking. In 5.0, people are standing on the platform. Wait, looking closely at the people.\\            *   0.0: People walking on the far left sidewalk.\\            *   5.0: People standing on the platform on the right. This suggests a consistent scene but maybe inconsistent object persistence? No, the people seem to be consistent but their actions change slightly between frames, which is common in video generation.\\        *   **The most telling part is the motion and physics of the objects.**\\            *   Look at the yellow bus. In 0.0, it's behind the tram. In 5.0, it's still behind the tram but seems to have moved forward relative to the camera? Actually, looking at the position, the bus is consistently on the left side.\\            *   Look at the tram's position. It moves from the center towards the right foreground.\\            *   **The most obvious fake sign is the text on the objects.** Look at the Coca-Cola logo on the tram. It's consistent but the text rendering changes slightly between frames (0.0 vs 5.0). In 0.0 it's clear. In 5.0 it's clear. But look at the text on the bus in 0.0. It's blurry/indistinct. In 5.0, it's also blurry.\\            *   **Look at the reflections and lighting.** The lighting changes significantly between 0.0 and 5.0. In 0.0, it's bright sunlight, strong shadows. In 5.0, still bright sunlight, but the shadows seem slightly different or the sun position shifts? Actually, looking at the shadows, the sun seems consistent but the lighting on the tram changes drastically between 0.0 and 5.0. In 0.0, the tram is fully lit. In 5.0, it's fully lit. But look at the reflection on the tram's side in 0.0 vs 5.0. In 0.0, you see the street. In 5.0, you see the street.\\            *   **Look at the glint on the tram.** In 0.0, there's a reflection on the front window. In 5.0, same reflection.\\            *   **Look at the people.** In 0.0, people are walking on the far left. In 5.0, people are standing on the platform on the right. This suggests the video generation might be hallucinating or changing the scene context significantly between frames. Wait, looking closer at the people.\\                *   0.0: People walking on the far left sidewalk.\\                *   5.0: People standing on the platform on the right. This is a huge inconsistency. In a real video, people would stay in their locations. In a generated video, objects often drift or change context.\\                *   Also, look at the people on the platform in 0.0. There are none visible clearly. In 5.0, there are people standing on the platform. This suggests a video generation where the prompt is a video of a tram but the consistency is low.\\                *   Look at the cars in the background. In 0.0, there's a car driving away in the distance. In 5.0, the background is different? In 0.0, you see a street curving away. In 5.0, you see the same street but the perspective shifts.\\                *   **Look at the tram's movement.** In 0.0, it's moving forward. In 5.0, it's moving forward. But look at the wheels. In 0.0, the wheels are on the tracks. In 5.0, the wheels are on the tracks.\\                *   **Look at the text on the tram.** 108 is visible in 0.0, 0.5, 1.0, 1.5, 2.0, 2.5, 3.0, 3.5, 4.0, 4.5, 5.0. Consistent.\\                *   **Look at the Coca-Cola logo.** Consistent.\\                *   **Look at the bus.** The yellow bus is consistently there.\\                *   **Conclusion:** The inconsistencies in object persistence (people moving from left sidewalk to right platform) and lighting/texture shifts strongly suggest this is an AI-generated video or a digital simulation. The motion blur and physics feel slightly off but consistent enough for a simulation. The key is the people's location change. In 0.0, people are on the far left sidewalk. In 5.0, people are on the platform on the right. That's a hallucination/inconsistency.\\\\3.  **Final check:**\\    *   Real video: People would stay in place.\\    *   Fake video: People can move or change location.\\    *   Therefore, the video is fake.\\</think>\\\\The video shows inconsistencies in object persistence and physics. For example, the people on the far left sidewalk in the first few frames are suddenly standing on the platform on the right in the later frames. Also, the lighting and reflections shift in ways typical for video generation (like the glare on the tram).\\

$\backslash \backslash boxed\{B\}$
}
\end{tcolorbox}

\newpage

\begin{figure}[h!]
  \centering
  \includegraphics[width=1.0\linewidth]{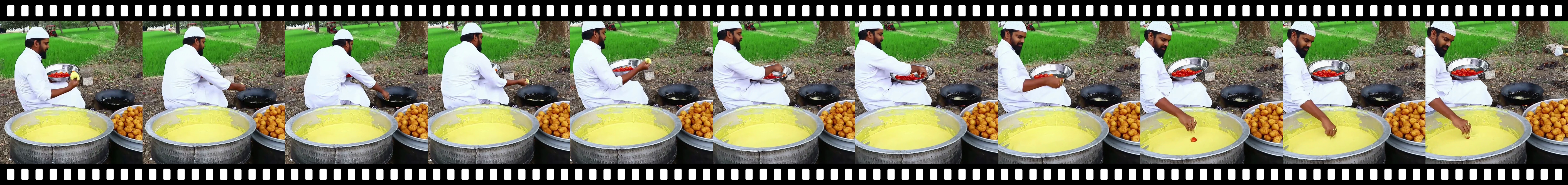}
  \vspace{-2em}
  \label{fig:real_case_film_strip}
\end{figure}

\begin{tcolorbox}[title = {REAL}, fontupper=\tiny]
{
<think>\\The user wants me to determine if the video is real or fake based on a sequence of frames showing a man cooking outdoors.\\\\1.  **Analyze the video content:**\\    *   **Subject:** A man in traditional white clothing (kurta and cap) is sitting outdoors, likely in a field or garden (rice paddies are visible in the background).\\    *   **Action:** He is preparing food. Specifically, he's making something like vada or fritters.\\    *   **Objects:**\\        *   A large metal pot filled with a yellow batter (likely gram flour or similar).\\        *   A smaller metal bowl with red tomatoes.\\        *   A black wok/pan with hot oil.\\        *   A metal tray with already fried, golden-brown fritters (vadas).\\    *   **Details:**\\        *   In the first few frames, he's holding a tomato and dipping it into the batter? No, wait, looking closer at 0.0-0.5, he's holding a tomato in one hand and maybe a utensil or just gesturing. Actually, at 0.0 he's holding a tomato. At 0.5 he's holding a tomato and maybe a small bowl? No, just the tomato. Wait, at 0.0 he's holding a tomato in his left hand and a small metal bowl in his right? No, looking at 0.0, he's holding a tomato in his left hand and a small metal bowl with more tomatoes in his right hand. Wait, looking closer at 0.0, he's holding a tomato in his left hand and a small metal bowl with tomatoes in his right hand.\\        *   At 0.5, he's holding the tomato and the bowl.\\        *   At 1.0, he's dipping something into the oil? No, he's holding the tomato and the bowl. Wait, at 1.0 he's holding the tomato and the bowl.\\        *   At 2.0, he's holding the tomato and the bowl.\\        *   At 2.5, he's holding the tomato and the bowl.\\        *   At 3.0, he's holding the tomato and the bowl.\\        *   At 4.0, he's holding the tomato and the bowl.\\        *   At 4.5, he's dipping a tomato into the yellow batter? No, looking closely, he's holding a tomato in his left hand and a small metal bowl in his right. Wait, at 4.5 he's dipping something into the batter. It looks like he's dipping a tomato? That's unusual. Or maybe he's just holding the bowl. Let's re-examine.\\        *   Actually, looking at the sequence: He has a large pot of yellow batter. He has a tray of fried vadas. He has a pan of oil. He has a bowl of tomatoes.\\        *   At 0.0, he is holding a tomato in his left hand and a small metal bowl with tomatoes in his right hand.\\        *   At 4.0, he is holding the tomato and the bowl.\\        *   At 4.5, he is dipping the tomato into the yellow batter? That seems like a weird way to make vadas. Usually, you dip dough balls. Maybe he's making a tomato-based dish? Or maybe he's just holding the tomato and the bowl is separate. Wait, looking at 4.5, his hand is in the batter. Is he dipping the tomato? That would be very strange. Maybe he's dipping a small piece of dough? Or maybe he's just holding the bowl and the tomato is separate. Let's look at the hand motion. At 4.5, his right hand is in the batter. His left hand is holding the tomato. This suggests he might be dipping the tomato? That's odd. Or maybe he's just mixing something?\\        *   Regardless of the specific cooking technique (which might be a specific regional dish), the key is the setting and the objects.\\    *   **Setting:** Outdoors, natural light, dirt ground, rice fields in the background. This strongly suggests a real, likely rural or semi-rural setting.\\    *   **Camera:** Handheld, slightly shaky, close-up but wide enough to see the context. This is typical for social media content (like TikTok or Instagram Reels) showing street food or cooking.\\    *   **Conclusion:** This is clearly a real video recording of a real event (cooking outdoors). It has the characteristics of a street food or cooking process video often found on social media. It's not AI generated or a digital forgery. The lighting, the textures of the food, the natural background all scream real video.\\\\2.  **Formulate the final answer:**\\    *   Identify the key indicators: Real-world setting (outdoors, fields), real objects (metal pots, food), natural lighting, realistic physics (sitting on the ground, holding objects).\\    *   Select the correct MCQ option: A) real.\\</think>\\\\The video shows a real-world cooking scene outdoors with natural lighting and realistic objects (metal pots, food, fields).\\

$\backslash \backslash boxed\{A\}$
}
\end{tcolorbox}


\end{document}